\documentclass[final]{cvpr}

\usepackage{times}
\usepackage{epsfig}
\usepackage{graphicx}
\usepackage{amsmath}
\usepackage{amssymb}

\usepackage[pagebackref=true,breaklinks=true,colorlinks,bookmarks=false]{hyperref}
\usepackage{multirow}
\newcommand{\xdownarrow}[1]{%
  {\left\downarrow\vbox to #1{}\right.\kern-\nulldelimiterspace}
}

\def\cvprPaperID{6805} 
\def\confYear{CVPR 2021}
\begin{document}

\title{Anomaly Detection in Video via Self-Supervised and Multi-Task Learning}

\author{Mariana-Iuliana Georgescu$^{1,3}$, Antonio B\u{a}rb\u{a}l\u{a}u$^{1}$, Radu Tudor Ionescu$^{1,3}$, Fahad Shahbaz Khan$^{2}$,\\
Marius Popescu$^{1,3}$, Mubarak Shah$^{4}$\\
$^1$University of Bucharest, Romania, $^2$MBZ University of Artificial Intelligence, Abu Dhabi\\
$^3$SecurifAI, Romania, $^4$University of Central Florida, Orlando, FL\vspace*{-0.3cm}
}

\maketitle

\begin{abstract}
\vspace{-0.1cm}
Anomaly detection in video is a challenging computer vision problem. Due to the lack of anomalous events at training time, anomaly detection requires the design of learning methods without full supervision. In this paper, we approach anomalous event detection in video through self-supervised and multi-task learning at the object level. We first utilize a pre-trained detector to detect objects. Then, we train a 3D convolutional neural network to produce discriminative anomaly-specific information by jointly learning multiple proxy tasks: three self-supervised and one based on knowledge distillation. The self-supervised tasks are: $(i)$ discrimination of forward/backward moving objects (arrow of time), $(ii)$ discrimination of objects in consecutive/intermittent frames (motion irregularity) and $(iii)$ reconstruction of object-specific appearance information. The knowledge distillation task takes into account both classification and detection information, generating large prediction discrepancies between teacher and student models when anomalies occur. To the best of our knowledge, we are the first to approach anomalous event detection in video as a multi-task learning problem, integrating multiple self-supervised and knowledge distillation proxy tasks in a single architecture. Our lightweight architecture outperforms the state-of-the-art methods on three benchmarks: Avenue, ShanghaiTech and UCSD Ped2. Additionally, we perform an ablation study demonstrating the importance of integrating self-supervised learning and normality-specific distillation in a multi-task learning setting.
\vspace{-0.1cm}
\end{abstract}

\setlength{\abovedisplayskip}{3.5pt}
\setlength{\belowdisplayskip}{3.5pt}

\vspace{-0.2cm}
\section{Introduction}
\vspace{-0.1cm}

In recent years, a growing interest has been dedicated to the task of detecting anomalous events in video \cite{Dong-Access-2020,Doshi-CVPRW-2020a,Doshi-CVPRW-2020b,Gong-ICCV-2019,Ionescu-CVPR-2019,Ionescu-WACV-2019,Ji-IJCNN-2020,Lee-TIP-2019,Lu-ECCV-2020,Nguyen-ICCV-2019,Pang-CVPR-2020,Park-CVPR-2020,Ramachandra-WACV-2020a,Ramachandra-WACV-2020b,Ramachandra-ArXiv-2020,Sun-ACMMM-2020,Tang-PRL-2020,Wang-ACMMM-2020,Wu-TNNLS-2019,Yu-ACMMM-2020,Zaheer-CVPR-2020,Zhang-PR-2020}. An anomalous event is commonly defined as an unfamiliar or unexpected event in a given context. For example, a person crossing the road can be viewed as anomalous if the event does not happen on the crosswalk. This example shows that context plays a key role in the definition of anomalous events and, consequently, in the problem formulation. Indeed, the reliance on context, coupled with the large variety of unexpected events, makes it extremely difficult to collect anomalous events for training. Hence, the anomaly detection problem is typically regarded as an outlier detection task. Then, a normality model is fit on normal training data, labeling events that deviate from the model as anomalous.
Without being able to employ standard supervision, researchers have proposed alternative approaches ranging from distance-based \cite{Ionescu-CVPR-2019,Ionescu-WACV-2019,Ramachandra-WACV-2020a,Ramachandra-WACV-2020b,Ravanbakhsh-WACV-2018,Sabokrou-IP-2017,Sabokrou-CVIU-2018,Saligrama-CVPR-2012,Smeureanu-ICIAP-2017,Sun-PR-2017,Tran-BMVC-2017,Xu-BMVC-2015} and reconstruction-based strategies \cite{Cong-CVPR-2011,Gong-ICCV-2019,Hasan-CVPR-2016,Liu-CVPR-2018,Lu-ICCV-2013,Luo-ICCV-2017,Nguyen-ICCV-2019,Park-CVPR-2020,Ravanbakhsh-ICIP-2017,Tang-PRL-2020,Vu-AAAI-2019} to probabilistic \cite{Adam-PAMI-2008,Antic-ICCV-2011,Cheng-CVPR-2015,Feng-NC-2017,Hinami-ICCV-2017,Kim-CVPR-2009,Mahadevan-CVPR-2010,Mehran-CVPR-2009,Wu-CVPR-2010} and change detection methods \cite{Giorno-ECCV-2016,Ionescu-ICCV-2017,Liu-BMVC-2018,Pang-CVPR-2020}. 

\begin{figure*}[!th]
\begin{center}
\includegraphics[width=1.0\linewidth]{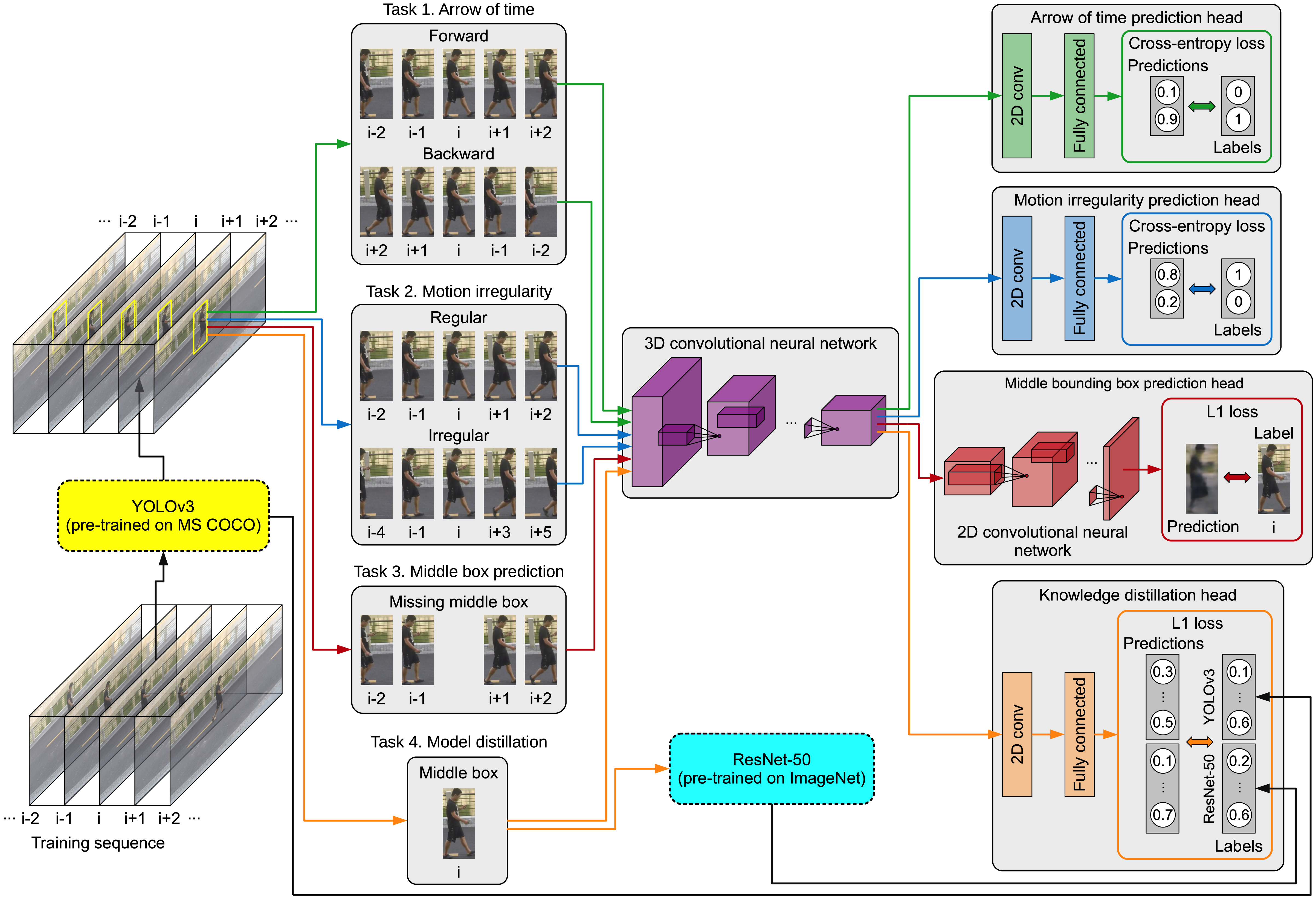}
\end{center}
\vspace{-0.35cm}
\caption{Our anomaly detection framework based on self-supervised and multi-task learning. First, we detect the objects in video with the help of an object detector (YOLOv3). For each object, we devise three self-supervised tasks (learning the arrow of time, predicting motion irregularity and predicting the object appearance in the middle box) and a knowledge distillation task (using YOLOv3 and ResNet-50 as teachers). A 3D convolutional neural network is trained jointly on the four tasks. Models represented with dashed lines are pre-trained. Best viewed in color.}
\label{fig:pipeline}
\vspace{-0.3cm}
\end{figure*}

In lieu of learning to discriminate directly between normal and anomalous events, related methods approach a different yet connected task. For example, in the pioneering work of Liu \etal~\cite{Liu-CVPR-2018}, a neural network learns to predict future video frames. During inference, an event is labeled as anomalous if the predicted future frame exhibits a high reconstruction error. Although the state-of-the-art methods attain impressive results, addressing anomaly detection through a single proxy task is suboptimal, since the proxy task is not well aligned with anomaly detection. For instance, a car stopped in a pedestrian area should be labeled as an anomaly, yet the car is trivial to reconstruct in a future frame (since it is standing still). We therefore propose to perform anomaly detection by training a model jointly on multiple proxy tasks. Following a series of recent methods \cite{Doshi-CVPRW-2020a,Doshi-CVPRW-2020b,Ionescu-CVPR-2019,Yu-ACMMM-2020}, we also employ an object detector, subsequently performing anomaly detection at the object level. However, these recent methods take into account a single proxy task. Different from \cite{Doshi-CVPRW-2020a,Doshi-CVPRW-2020b,Ionescu-CVPR-2019,Yu-ACMMM-2020}, we propose a novel anomaly detection approach that jointly learns a set of multiple proxy tasks through a single object-centric architecture.  

As discussed above, we devise an object-centric approach comprising a 3D convolutional neural network (CNN) that jointly learns the following proxy tasks: $(i)$ predicting the arrow of time (discriminating between forward and backward moving objects), $(ii)$ predicting the irregularity of motion (discriminating between objects captured in consecutive frames versus objects captured in intermittent frames), $(iii)$ reconstructing the appearance of objects (given their appearance in preceding and succeeding frames), $(iv)$ estimating normality-specific class probabilities by distilling pre-trained classification (ImageNet \cite{Russakovsky2015}) and detection (MS COCO \cite{Lin-ECCV-2014}) teachers. 
To jointly address these self-supervised and knowledge distillation tasks, we integrate a prediction head for each corresponding task, as illustrated in Figure~\ref{fig:pipeline}. To our knowledge, we are the first to propose a multi-task learning approach that integrates a set of novel self-supervised and knowledge distillation proxy tasks in a single object-centric architecture for anomaly detection in video.

We perform comprehensive experiments on three benchmarks, namely  Avenue \cite{Lu-ICCV-2013}, ShanghaiTech \cite{Luo-ICCV-2017} and UCSD Ped2 \cite{Mahadevan-CVPR-2010}. Our approach outperforms the state-of-the-art methods \cite{Giorno-ECCV-2016,Dong-Access-2020,Doshi-CVPRW-2020a,Doshi-CVPRW-2020b,Gong-ICCV-2019,Hasan-CVPR-2016,Hinami-ICCV-2017,Ionescu-CVPR-2019,Ionescu-ICCV-2017,Ionescu-WACV-2019,Ji-IJCNN-2020,Kim-CVPR-2009,Lee-ICASSP-2018,Lee-TIP-2019,Liu-CVPR-2018,Liu-BMVC-2018,Lu-ICCV-2013,Lu-ECCV-2020,Luo-ICCV-2017,Mahadevan-CVPR-2010,Mehran-CVPR-2009,Nguyen-ICCV-2019,Park-CVPR-2020,Ramachandra-WACV-2020a,Ramachandra-WACV-2020b,Ravanbakhsh-WACV-2018,Ravanbakhsh-ICIP-2017,Smeureanu-ICIAP-2017,Sultani-CVPR-2018,Sun-ACMMM-2020,Tang-PRL-2020,Vu-AAAI-2019,Wang-ACMMM-2020,Wu-TNNLS-2019,Xu-BMVC-2015,Xu-CVIU-2017,Yu-ACMMM-2020,Zaheer-CVPR-2020,Zhang-PR-2016} on all three data sets, achieving frame-level AUC scores of $92.8\%$ on Avenue, $90.2\%$ on ShanghaiTech and $99.8\%$ on UCSD Ped2. 
Additionally, we present empirical evidence confirming that a jointly optimized model on the proposed proxy tasks outperforms single models optimized on individual tasks, thus indicating that modeling anomaly detection through a single proxy task is suboptimal.

In summary, our contribution is multifold:
\begin{itemize}
\vspace{-0.25cm}
    \item We introduce learning the arrow of time as a proxy task for anomaly detection.
\vspace{-0.3cm}
    \item We introduce motion irregularity prediction as a proxy task for anomaly detection.
\vspace{-0.3cm}
    \item We introduce model distillation as a proxy task for anomaly detection in video.
\vspace{-0.3cm}
    \item We pose anomaly detection in video as a multi-task learning problem, integrating multiple self-supervised and knowledge distillation tasks into a \emph{single} model.
\vspace{-0.3cm}
    \item We conduct experiments showing that our approach attains superior results compared to the state-of-the-art methods on three benchmarks.
\vspace{-0.3cm}
\end{itemize}


\section{Related Work}
\label{sec_related}
\vspace{-0.1cm}

While the early works \cite{Adam-PAMI-2008,Antic-ICCV-2011,Cong-2013,Li-PAMI-2014,Lu-ICCV-2013,Mahadevan-CVPR-2010,Mehran-CVPR-2009,Saligrama-CVPR-2012,Wu-CVPR-2010} on video anomaly detection relied heavily on handcrafted appearance and motion features, the recent literature is abundant in deep learning methods \cite{Doshi-CVPRW-2020a,Doshi-CVPRW-2020b,Hasan-CVPR-2016,Hinami-ICCV-2017,Ionescu-CVPR-2019,Liu-CVPR-2018,Luo-ICCV-2017,Ramachandra-WACV-2020b,Ravanbakhsh-WACV-2018,Ravanbakhsh-ICIP-2017,Sabokrou-IP-2017,Smeureanu-ICIAP-2017,Wang-ICIP-2018,Xu-BMVC-2015,Xu-CVIU-2017}. 
For instance, Xu \etal~\cite{Xu-BMVC-2015} proposed the use of stacked denoising auto-encoders to automatically learn both appearance and motion features, which are further used as input for multiple one-class SVM models.
Hasan \etal~\cite{Hasan-CVPR-2016} diverged from using auto-encoders simply as feature extractors for subsequent models, leveraging the reconstruction error as an estimator for abnormality. More recently, Wang \etal~\cite{Wang-ICIP-2018} proposed a further improvement by combining CNNs with LSTMs, forming a spatio-temporal auto-encoder able to better account for the temporal evolution of spatial features. Wang \etal~\cite{Wang-ICIP-2018} rely on the assumption that anomalous events will cause significant discrepancies between future and past frames. 
Employing generative networks for video anomaly detection \cite{Dong-Access-2020,Park-CVPR-2020,Ravanbakhsh-ICIP-2017} is another significant line of research that relies on the same principle, that is, synthesizing future frames will prove to be significantly more challenging when an anomalous event occurs than in a normal situation. To this end, Liu \etal~\cite{Liu-CVPR-2018} employed a generative model to predict future frames, considering the reconstruction error as an indicator of abnormality. 
In another similar framework, Lee \etal~\cite{Lee-TIP-2019} proposed to predict the middle frame, considering a bidirectional approach that learns from both past and future frames. Similar to future frame \cite{Dong-Access-2020,Liu-CVPR-2018} or middle frame \cite{Lee-TIP-2019} prediction frameworks, we propose a framework that incorporates middle frame prediction. Different from methods such as \cite{Dong-Access-2020,Lee-TIP-2019,Liu-CVPR-2018,Wang-ICIP-2018}, we study middle frame prediction at the object level, enabling the accurate localization of anomalies. Moreover, middle frame prediction is just one of our four proxy tasks. To our knowledge, we are the first to propose learning the arrow of time, motion irregularity prediction and model distillation as proxy tasks for anomaly detection in video. We note that model distillation has been studied as a single task for anomaly detection in still images~\cite{Bergmann-CVPR-2020}. However, our ablation results show that model distillation alone is not sufficient for anomaly detection in video.

Aside from the direction relying on reconstruction errors \cite{Hasan-CVPR-2016,Liu-CVPR-2018,Lu-ICCV-2013,Luo-ICCV-2017,Nguyen-ICCV-2019,Park-CVPR-2020,Ravanbakhsh-ICIP-2017,Tang-PRL-2020,Vu-AAAI-2019}, other recent works, such as \cite{Doshi-CVPRW-2020a,Ramachandra-WACV-2020b}, tackle the problem from completely different angles. For example, Ramachandra \etal~\cite{Ramachandra-WACV-2020b} employed a Siamese network to learn a metric between spatio-temporal video patches. 
In this scenario, the dissimilarity between patches provides the means to estimate the level of abnormality. 

In addition, anomalous event detection approaches can be divided with respect to the level of analysis. While some frameworks, such as \cite{Liu-CVPR-2018,Mehran-CVPR-2009,Ravanbakhsh-WACV-2018,Ravanbakhsh-ICIP-2017,Smeureanu-ICIAP-2017}, approach the problem from a global (frame-level) perspective, methods such as \cite{Giorno-ECCV-2016,Dutta-AAAI-2015,Kim-CVPR-2009,Ionescu-WACV-2019,Liu-BMVC-2018,Lu-ICCV-2013,Luo-ICCV-2017,Mahadevan-CVPR-2010,Sabokrou-IP-2017,Saligrama-CVPR-2012,Zhang-PR-2016} extract features at a local level, \eg by considering spatio-temporal cubes. In some cases, the detection of anomalous events is explored with multi-level frameworks, a recent example being the work of Lee \etal~\cite{Lee-TIP-2019}. Aside from these mainstream perspectives, Ionescu \etal~\cite{Ionescu-CVPR-2019} introduced a novel object-centric framework, employing a single-shot object detector on each frame, before applying convolutional auto-encoders to learn deep unsupervised representations as part of a one-versus-rest classification approach based on clustering training samples into normality clusters. A few recent works \cite{Doshi-CVPRW-2020a,Doshi-CVPRW-2020b,Yu-ACMMM-2020} further explored the same line of research, proposing alternative object-centric frameworks. Similar to object-centric frameworks such as \cite{Doshi-CVPRW-2020a,Doshi-CVPRW-2020b,Ionescu-CVPR-2019,Yu-ACMMM-2020}, we employ an object detector, focusing our analysis on the detected objects. Unlike \cite{Doshi-CVPRW-2020a,Doshi-CVPRW-2020b,Ionescu-CVPR-2019,Yu-ACMMM-2020}, we perform the analysis through a series of proxy self-supervised and model distillation tasks, proposing a novel anomaly detection framework based on multi-task learning. Hence, the only common aspect with the other object-centric methods \cite{Doshi-CVPRW-2020a,Doshi-CVPRW-2020b,Ionescu-CVPR-2019,Yu-ACMMM-2020} is the use of an object detector.

The related methods presented so far follow the mainstream formulation of anomalous event detection, which implies that an anomalous event is an unfamiliar event in a known context. In the mainstream formulation, anomalous events are not available at training time, as it is considered too difficult to collect a sufficiently wide variety of anomalous events. Although our study adopts the mainstream formulation, we acknowledge the recent effort of Sultani \etal~\cite{Sultani-CVPR-2018}, which considers anomalous events that do not depend on the context. 
By eliminating the reliance on context, they are able to collect and use anomalous events at training. In their formulation, anomalous event detection becomes equivalent to action recognition in video. We thus consider the line of research initiated by Sultani \etal~\cite{Sultani-CVPR-2018} and continued by others~\cite{Zhong-CVPR-2019} less related to our study.

\vspace*{-0.1cm}
\section{Method}
\label{sec_method}
\vspace{-0.1cm}
\subsection{Motivation and Overview}
\vspace*{-0.1cm}

\noindent
{\bf Motivation.}
Modeling anomalous event detection through a single proxy task, \eg future frame prediction \cite{Liu-CVPR-2018}, is suboptimal due to the lack of perfect alignment between the proxy task and the actual (anomaly detection) task. To reduce the non-alignment of the model with respect to the anomaly detection task, we propose to train the model by jointly optimizing it on multiple proxy tasks.

\noindent
{\bf Training.}
Our framework based on self-supervised and multi-task learning is illustrated in Figure \ref{fig:pipeline}. First, we detect the objects in each frame using a pre-trained YOLOv3 \cite{Redmon-arXiv-2018} detector, obtaining a list of bounding boxes. For each detected object in the frame $i$, we create an \emph{object-centric temporal sequence} by simply cropping the corresponding bounding box from frames $\{i-t, ..., i-1, i, i+1, ...., i+t \}$ (without performing any object tracking), resizing each cropped image to $64\times64$ pixels. For illustration purposes, we set $t=2$ in Figure \ref{fig:pipeline}. The resulting object-centric sequence is the input of our 3D CNN. Our architecture is formed of the shared 3D CNN followed by four branches (prediction heads), one for each proxy task. 

\noindent
{\bf Inference.}
During inference, the anomaly score is computed by averaging the scores predicted for each task. For the arrow of time and motion irregularity tasks, we take the probability of the temporal sequence to move backward and the probability of the temporal sequence to be intermittent. For the middle frame prediction task, we consider the mean absolute difference between the ground-truth and the reconstructed object. The last component of the anomaly score is the difference between the class probabilities predicted by YOLOv3 and the corresponding class probabilities predicted by our knowledge distillation branch. We do not include ResNet-50 predictions at inference time to preserve the real-time processing of our framework. 

\vspace*{-0.1cm}
\subsection{Neural Architectures}  
\vspace*{-0.1cm}

\begin{table}[t] 
\small{
\begin{center}
\begin{tabular}{|l c|c|c|}
\hline 
 \multicolumn{3}{|c|}{}\vspace{-0.3cm}\\
 \multicolumn{3}{|c|}{\vspace{-0.5cm}$\xrightarrow{\hspace*{3.1cm}{\mbox{width}}\hspace*{3.1cm}}$}   \\
 \multirow{1}{*}{\rotatebox{90}{$\xleftarrow{\hspace*{2.85cm}{\mbox{depth}}\hspace*{2.85cm}}$}} \\
  & $3 \times 3 \times 3$ conv $16$   & $3 \times 3 \times 3$ conv $32$ \\
  & $1 \times 2 \times 2$ max-pooling & $1 \times 2 \times 2$ max-pooling \\
  & $3 \times 3 \times 3$ conv $32$   & $3 \times 3 \times 3$ conv $64$ \\
  & $1 \times 2 \times 2$ max-pooling & $1 \times 2 \times 2$ max-pooling \\
  & $3 \times 3 \times 3$ conv $32$   & $3 \times 3 \times 3$ conv $64$ \\
  & $: \times 2 \times 2$ max-pooling & $: \times 2 \times 2$ max-pooling \\ 
  & & \vspace{-0.2cm}\\
 \cline{2-3}
   & & \vspace{-0.2cm}\\
  & $3 \times 3 \times 3$ conv $16$  & $3 \times 3 \times 3$ conv $32$ \\
  & $3 \times 3 \times 3$ conv $16$  & $3 \times 3 \times 3$ conv $32$ \\
  & $1 \times 2 \times 2$ max-pooling & $1 \times 2 \times 2$ max-pooling \\
  & $3 \times 3 \times 3$ conv $32$  & $3 \times 3 \times 3$ conv $64$ \\
  & $3 \times 3 \times 3$ conv $32$  & $3 \times 3 \times 3$ conv $64$ \\
  & $1 \times 2 \times 2$ max-pooling & $1 \times 2 \times 2$ max-pooling \\
  & $3 \times 3 \times 3$ conv $32$  & $3 \times 3 \times 3$ conv $64$ \\
  & $1 \times 2 \times 2$ max-pooling & $1 \times 2 \times 2$ max-pooling \\
  & $3 \times 3 \times 3$ conv $32$  & $3 \times 3 \times 3$ conv $64$ \\
  & $: \times 2 \times 2$ max-pooling & $: \times 2 \times 2$ max-pooling \\
\hline
\end{tabular}
\end{center}
\vspace*{-0.2cm}
}
\caption{Alternative architectures considered for the 3D CNN included in our anomaly detection framework. Global temporal pooling is denoted by ``:''.}\label{table:networks_arch}
\vspace*{-0.25cm}
\end{table}

Our architecture is composed of a shared CNN and four independent prediction heads. The shared CNN uses 3D convolutions (conv) to model temporal dependencies, while individual branches use only 2D convolutions. When considering the proxy tasks one at a time, we observed accurate results using a relatively shallow and narrow neural architecture formed of three conv layers. When we moved to jointly optimizing our model on multiple proxy tasks, we observed the need to increase the width and depth of our neural network to accommodate for the increased complexity of the multi-task learning problem. We therefore employ a set of four neural architectures considering all possible combinations of shallow, deep, narrow and wide architectures. These are: shallow+narrow, shallow+wide, deep+narrow and deep+wide. The detailed configuration of each 3D CNN architecture is presented in Table \ref{table:networks_arch}. 

For each network configuration, the spatial size of the RGB input is $64 \times 64$ pixels. The 3D conv layers use filters of $3 \times 3 \times 3$. Each conv layer is followed by a batch normalization layer and a ReLU activation. Our shallow+narrow 3D CNN is formed of three 3D conv layers and three 3D max-pooling layers. Its first 3D conv layer is composed of $16$ filters and the next two conv layers are composed of $32$ filters each. The padding is set to ``same'' and the stride is set to $1$. We perform only spatial pooling for the first two 3D max-pooling layers. The pooling size and the stride are both set to $2$. The last 3D max-pooling layer performs global temporal pooling, keeping the same configuration as the first two pooling layers at the spatial level. Using temporal pooling only once (in the last pooling layer) enables us to employ a different temporal size for each proxy task. In the shallow+wide configuration, we change the 3D CNN by doubling the number of filters in each conv layer. For the deep+narrow architecture, we increase the number of 3D conv layers from three to six. Finally, in the deep+wide configuration, we double the number of layers as well as the number of filters in each conv layer with respect to the shallow+narrow model.

In the middle object prediction head, we incorporate a decoder formed of upsampling and 2D conv layers based on $3\times3$ filters. The number of upsampling operations is always equal to the number of max-pooling layers in the 3D CNN. Similarly, the number of 2D conv layers in the decoder matches the number of 3D conv layers in the 3D CNN. Each upsampling operation is based on nearest neighbor interpolation, increasing the spatial support by a factor of $2\times$. The last conv layer in the decoder has only three filters in order to reconstruct the RGB input. 

The other three prediction heads share the same configuration, having a 2D conv layer with $32$ filters and a max-pooling layer with a spatial support of $2 \times 2$. The last layer is a fully-connected layer with either two units to predict the arrow of time and motion irregularity or $1080$ units to predict the teachers' output scores for the $1000$ ImageNet \cite{Russakovsky2015} classes and the $80$ MS COCO \cite{Lin-ECCV-2014} categories.

\vspace*{-0.1cm}
\subsection{Proxy Tasks and Joint Learning}
\vspace*{-0.1cm}

\noindent
{\bf Task 1: Arrow of time.}
To predict the arrow of time \cite{Wei-CVPR-2018} at the object level, we generate two labeled training samples from each object-centric sequence. The first sample takes the frames in their temporal order, namely $(i-t, ..., i-1, i, i+1,..., i+t)$, thus being labeled as forward motion (class $1$). The second sample takes the frames in reversed order, namely $(i+t, ..., i+1, i, i-1,..., i-t)$, being labeled as backward motion (class $2$). During inference, we expect the arrow of time to be harder to predict for objects with anomalous motion.
Let $f$ be the shared 3D CNN and $h_{T_1}$ be the arrow of time head. Let $X^{(T_1)}$ be a forward or backward object-centric sequence of size $(2\cdot t+1) \times 64 \times 64 \times 3$. We use the cross-entropy loss to train the arrow of time head:
\begin{equation}\label{eq_arrow_of_time}
\mathcal{L}_{T_1} \left( X^{(T_1)}, Y^{(T_1)}\right) = -\sum_{k=1}^{2} Y^{(T_1)}_k log\left(\hat{Y}^{(T_1)}_k\right) ,
\end{equation}
where $\hat{Y}^{(T_1)}\!=\!softmax \left( h_{T_1}\left(f(X^{(T_1)}\right) \right)$ and $Y^{(T_1)}$ is the one-hot encoding of the ground-truth label for $X^{(T_1)}$.
 
\noindent
{\bf Task 2: Motion irregularity.}
Assuming that some anomalies can be identified through irregular motion patterns, we train our model to predict if an object-centric sequence has consecutive or intermittent frames (some frames being skipped). To learn motion irregularity, we generate two labeled training samples from each object-centric sequence. The first example captures an object in consecutive frames from $i-t$ to $i+t$, the corresponding class being regular motion (class $1$). The intermittent object-centric sequence is created by retaining the frame $i$, then gradually adding $t$ randomly chosen previous frames and $t$ randomly chosen succeeding frames. The intermittent frames are chosen by skipping frames using random gaps in the range $\{1, 2, 3, 4 \}$. The intermittent object-centric sequence is labeled as irregular motion (class $2$). Let $h_{T_2}$ be the irregular motion head and $X^{(T_2)}$ be a regular or irregular object-centric sequence of size $(2\cdot t+1) \times 64 \times 64 \times 3$. We employ the cross-entropy loss to train the motion irregularity head:
\begin{equation}\label{eq_irregular_motion}
\mathcal{L}_{T_2} \left(X^{(T_2)}, Y^{(T_2)}\right) = -\sum_{k=1}^{2}Y^{(T_2)}_k log\left(\hat{Y}^{(T_2)}_k\right),
\end{equation}
where $\hat{Y}^{(T_2)}\!=\!softmax\left(h_{T_2} \left(f(X^{(T_2)}\right)\right)$ and $Y^{(T_2)}$ is the one-hot encoding of the ground-truth label for $X^{(T_2)}$.

\noindent
{\bf Task 3: Middle bounding box prediction.}
Our 3D CNN model also learns to reconstruct objects detected in the normal training videos. 
From each object-centric sequence, we select the image samples cropped from frames $\{i-t, ..., i-1, i+1,..., i+t\}$, forming the input object-centric sequence $X^{(T_3)}$ of size $(2\cdot t) \times 64 \times 64 \times 3$. The middle image, corresponding to the bounding box in frame $i$, represents the target output $Y^{(T_3)}$ of size $64 \times 64 \times 3$. When we encounter an anomaly with unusual motion, such as a person running, the input object-centric sequence of that person will not contain enough information for the model to accurately reconstruct the middle bounding box, thus being labeled as anomalous. Let $h_{T_3}$ be the middle bounding box prediction head. We use the $L_1$ loss to learn the middle bounding box prediction task:
\begin{equation}\label{eq_middle_box}
\mathcal{L}_{T_3}\!\! \left(\!X^{(T_3)}, \!Y^{(T_3)\!}\right)\!\!=\!\!\frac{1}{h\!\cdot\!w\!\cdot\!c}\!
  \sum_{j=1}^{h}\!\sum_{k=1}^{w}\!\sum_{l=1}^{c} \! \left| Y^{(T_3)}_{jkl} \!-\!\hat{Y}^{(T_3)}_{jkl}\right|\!,\!
\end{equation}
where $\hat{Y}^{(T_3)}\!=\!h_{T_3}\left(f\left(X^{(T_3)}\right)\right)$ and $h\times w \times c$ is the size of the output, \ie $h=64$, $w=64$ and $c=3$.

\noindent
{\bf Task 4: Model distillation.}
On the one hand, our 3D CNN model learns to predict the features from the last layer (just before softmax) of a ResNet-50 \cite{He-CVPR-2016}, which is pre-trained on ImageNet. On the other hand, our 3D CNN model learns to predict the class probabilities predicted by YOLOv3 \cite{Redmon-arXiv-2018}, which is pre-trained on MS COCO. During distillation, our model learns the predictive behavior of the teachers on normal data. During inference, we expect high prediction discrepancies between our student and the YOLOv3 teacher when we encounter an object with unusual appearance or that belongs to an object category not seen during training. We refrain from using ResNet-50 during inference in order to save valuable computational time.
We note that YOLOv3 is applied only once on each frame $i$, the corresponding class probabilities for each detected object being already available during model distillation. During training, we still need to pass each object to ResNet-50 to extract the pre-softmax features. In order to distill the knowledge from the YOLOv3 and ResNet-50 teachers, our student 3D CNN model receives the same input as ResNet-50 and learns to predict the pre-softmax features $Y^{(T_4)}_{\mbox{\scriptsize{ResNet}}}$ of ResNet-50 and the class probabilities $Y^{(T_4)}_{\mbox{\scriptsize{YOLO}}}$ predicted by YOLOv3. Let $X^{(T_4)}$ be the input image comprising a detected object and $h_{T_4}$ be the knowledge distillation head. The model distillation task is learned by minimizing the $L_1$ loss function:
\begin{equation}\label{eq_model_distillation}
\mathcal{L}_{T_4} \left(X^{(T_4)}, Y^{(T_4)}\right) = \frac{1}{n}
  \sum_{j=1}^{n}\left|Y^{(T_4)}_j - \hat{Y}^{(T_4)}_j\right|,
\end{equation}
where $\hat{Y}^{(T_4)}\!=\!h_{T_4}\!\left(f\!\left(X^{(T_4)}\right)\!\right)$ and $Y^{(T_4)}\!=\!Y^{(T_4)}_{\mbox{\scriptsize{ResNet}}} \oplus Y^{(T_4)}_{\mbox{\scriptsize{YOLO}}}$ is the concatenation of the $1000$ ResNet-50 pre-softmax features and the $80$ YOLOv3 class probabilities, resulting in a vector of $n=1080$ components. 

\noindent
{\bf Joint loss.}
Our 3D CNN model is trained by jointly optimizing it on the four proxy tasks described above. Hence, the model is training using a joint loss function:
\begin{equation}\label{eq_total_loss}
\mathcal{L}_{\mbox{\scriptsize{total}}} = \mathcal{L}_{T_1} + \mathcal{L}_{T_2} + \mathcal{L}_{T_3} + \lambda \cdot \mathcal{L}_{T_4},
\end{equation}
where $\lambda \in (0, 1]$ is a weight that regulates the importance of the knowledge distillation task. We empirically observed that $\mathcal{L}_{T_4}$ has a typically higher magnitude than the other loss functions, dominating the joint loss without a regularization term. In our experiments, we fine-tune $\lambda$ with respect to the validation values of the joint loss, before ever applying our framework on the anomaly detection task.


\vspace*{-0.1cm}
\subsection{Inference}
\vspace*{-0.1cm}

During inference, we utilize YOLOv3 to detect objects in each frame $i$. For each object, we extract the corresponding object-centric sequence $X$ by cropping the bounding box from the frames $\{i-t,..., i-1, i, i+1,..., i+t \}$. 
We pass each object-centric sequence through our neural model, obtaining the outputs $\hat{Y}^{(T_1)}$, $\hat{Y}^{(T_2)}$, $\hat{Y}^{(T_3)}$ and $\hat{Y}^{(T_4)}$, respectively. For the arrow of time proxy task, we take the probability of the temporal sequence to move backward as the anomaly score. For the motion irregularity task, we consider the probability of the gapless test sequence $X$ to be intermittent as a good abnormality indicator. We interpret the mean absolute error between the reconstructed and the ground-truth middle object as the anomaly score provided by the middle bounding box prediction head. For the knowledge distillation task, we consider the absolute difference between the class probabilities predicted by YOLOv3 and those predicted by our model. We compute the final anomaly score of an object as the average of the anomaly scores given by each prediction head:
\begin{equation}\label{eq_total}
\begin{split}
s&core(X) = \frac{1}{4}\Big(\hat{Y}^{(T_1)}_2 + \hat{Y}^{(T_2)}_2 + \\
& avg\!\left(\left|Y^{(T_3)}\!-\!\hat{Y}^{(T_3)}\right|\right) + avg\!\left(\left|Y^{(T_4)}_{\mbox{\scriptsize{YOLO}}}\!-\! \hat{Y}^{(T_4)}_{\mbox{\scriptsize{YOLO}}} \right| \right)\!\!\Big) .
\end{split}
\end{equation}
Next, we reassemble the detected objects in a pixel-level anomaly map for each frame. Therefore, we can easily localize the anomalous regions in any given frame. To create a smooth pixel-level anomaly map, we apply a 3D mean filter. The anomaly score for a certain frame is given by the maximum score in the corresponding anomaly map. The final frame-level anomaly scores are obtained by applying a temporal Gaussian filter.

\vspace*{-0.1cm}
\subsection{Object-Level versus Frame-Level Detection}
\vspace*{-0.1cm}

Although performing anomaly detection at the object level enables the accurate localization of anomalies, the downside is that the detection failures of YOLOv3 (due to a limited set of object categories or poor performance) are translated into false negatives. In order to address this limitation, we can apply our framework at the frame level, eliminating YOLOv3 from the pipeline and keeping the other components in place. By fusing the frame-level and object-level anomaly scores at a late stage, we can recover some of the false negatives of our object-centric framework. In our experiments, we report the results of our framework based on late fusion, as well as the results at the object level and at the frame level, respectively.

\vspace*{-0.1cm}
\section{Experiments}
\label{sec_experiments}
\vspace{-0.1cm}
 
\subsection{Data Sets}
\vspace*{-0.1cm}

We perform experiments on three benchmark data sets: Avenue \cite{Lu-ICCV-2013}, ShanghaiTech \cite{Luo-ICCV-2017} and UCSD Ped2 \cite{Mahadevan-CVPR-2010}. Each data set has pre-defined training and test sets, anomalous events being included only at test time. 

\noindent
{\bf Avenue.} The Avenue \cite{Lu-ICCV-2013} data set contains 16 training videos with normal activity and 21 test videos. Examples of anomalous events in Avenue are related to people running, throwing objects or walking in the wrong direction. The resolution of each video is $360 \times 640$ pixels. 

\noindent
{\bf ShanghaiTech.} ShanghaiTech \cite{Luo-ICCV-2017} is one of the largest data sets for anomaly detection in video. It consists of $330$ training videos and $107$ test videos. The training videos contain only normal events, while the test videos contain normal and abnormal sequences. Examples of anomalous events are: robbing, jumping, fighting and riding bikes in pedestrian areas. The resolution of each video is $480\times856$ pixels. 
 
\noindent
{\bf UCSD Ped2.} UCSD Ped2 \cite{Mahadevan-CVPR-2010} contains 16 training videos with normal activity and 12 test videos. Examples of abnormal events are bikers, skaters and cars in a pedestrian area. The resolution of each video is $240\times360$ pixels. 

\vspace*{-0.1cm}
\subsection{Setup and Implementation Details}
\vspace*{-0.1cm}

\noindent
{\bf Evaluation measures.}
As our main evaluation metric, we consider the area under the curve (AUC) computed with respect to the ground-truth frame-level annotations. The frame-level AUC metric is the most commonly used metric in related works \cite{Giorno-ECCV-2016,Gong-ICCV-2019,Hasan-CVPR-2016,Hinami-ICCV-2017,Ionescu-CVPR-2019,Liu-CVPR-2018,Ramachandra-ArXiv-2020,Ravanbakhsh-ICIP-2017,Vu-AAAI-2019,Wang-ACMMM-2020,Zaheer-CVPR-2020}. Many related works also report the pixel-level AUC for the UCSD Ped2 data set. As explained by Ramachandra \etal~\cite{Ramachandra-WACV-2020a}, the pixel-level AUC is a flawed evaluation metric. We thus report our performance on UCSD Ped2 in terms of the region-based detection criterion (RBDC) and the track-based detection criterion (TBDC). These metrics were recently introduced by Ramachandra \etal~\cite{Ramachandra-WACV-2020a} to replace the commonly used pixel-level and frame-level AUC metrics.


\begin{table}[t]
\setlength\tabcolsep{4.0pt}
\small{
\begin{center}
\begin{tabular}{|c|l|c|c|c|c|c|c|c|c|c|c|c|c|c|c|}
\hline
\multirow{2}{*}{Year} & \multirow{2}{*}{Method} & \multirow{2}{*}{Avenue} & Shanghai & UCSD \\
& &  & Tech & Ped2 \\
\hline
\multirow{5}{*}{\rotatebox{90}{before 2016\hspace{0.1cm}}}&Kim \etal~\cite{Kim-CVPR-2009} & - & - & $69.3$ \\
&Mehran \etal~\cite{Mehran-CVPR-2009} & - & - & $55.6$ \\
&Mahadevan \etal~\cite{Mahadevan-CVPR-2010} & - & - & $82.9$ \\
&Lu \etal~\cite{Lu-ICCV-2013} & $80.9$ & - & - \\
&Xu \etal~\cite{Xu-BMVC-2015} & - & - & $90.8$ \\
\hline
\multirow{3}{*}{\rotatebox{90}{2016\hspace{0.1cm}}}&Del Giorno \etal~\cite{Giorno-ECCV-2016} & $78.3$ & - & - \\
&Hasan \etal~\cite{Hasan-CVPR-2016} & $70.2$ & $60.9$ & $90.0$ \\
&Zhang \etal~\cite{Zhang-PR-2016} & - & - & $91.0$ \\
\hline
\multirow{6}{*}{\rotatebox{90}{2017\hspace{0.1cm}}}&Hinami \etal~\cite{Hinami-ICCV-2017} & - & - & $92.2$ \\
&Ionescu \etal~\cite{Ionescu-ICCV-2017} & $80.6$ & - & $82.2$ \\
&Luo \etal~\cite{Luo-ICCV-2017} & $81.7$ & $68.0$ & $92.2$ \\
&Ravanbakhsh \etal~\cite{Ravanbakhsh-ICIP-2017} & - & - & $93.5$ \\
&Smeureanu \etal~\cite{Smeureanu-ICIAP-2017} & $84.6$ & - & - \\
&Xu \etal~\cite{Xu-CVIU-2017} & - & - & $90.8$ \\
\hline
\multirow{5}{*}{\rotatebox{90}{2018\hspace{0.1cm}}}&Lee \etal~\cite{Lee-ICASSP-2018} & $87.2$ & - & $96.5$ \\
&Liu \etal~\cite{Liu-CVPR-2018} & $85.1$ & $72.8$ & $95.4$ \\
&Liu \etal~\cite{Liu-BMVC-2018} & $84.4$ & - & $87.5$ \\
&Ravanbakhsh \etal~\cite{Ravanbakhsh-WACV-2018} & - & - & $88.4$ \\
&Sultani \etal~\cite{Sultani-CVPR-2018} & - & $76.5$ & - \\
\hline
\multirow{7}{*}{\rotatebox{90}{2019\hspace{0.1cm}}}&Gong \etal~\cite{Gong-ICCV-2019} & $83.3$ & $71.2$ & $94.1$ \\
&Ionescu \etal~\cite{Ionescu-CVPR-2019} & $90.4$ & $84.9$ & $97.8$ \\
&Ionescu \etal~\cite{Ionescu-WACV-2019} & $88.9$ & - & - \\
&Lee \etal~\cite{Lee-TIP-2019} & $90.0$ & $76.2$ & $96.6$ \\
&Nguyen \etal~\cite{Nguyen-ICCV-2019} & $86.9$ & - & $96.2$ \\
&Vu \etal~\cite{Vu-AAAI-2019} & $71.5$ & - & $\color{blue}{99.2}$ \\
&Wu \etal~\cite{Wu-TNNLS-2019} & $86.6$ & - & - \\
\hline
\multirow{15}{*}{\rotatebox{90}{2020\hspace{0.1cm}}}
&Dong \etal~\cite{Dong-Access-2020} & $84.9$ & $73.7$ & $95.6$ \\
&Doshi \etal~\cite{Doshi-CVPRW-2020a,Doshi-CVPRW-2020b} & $86.4$ & $71.6$ & $97.8$ \\
&Ji \etal~\cite{Ji-IJCNN-2020} & $78.3$ & - & $98.1$ \\
&Lu \etal~\cite{Lu-ECCV-2020} & $85.8$ & $77.9$ & $96.2$ \\
&Park \etal~\cite{Park-CVPR-2020} & $88.5$ & $70.5$ & $97.0$ \\
&Ramachandra \etal~\cite{Ramachandra-WACV-2020a} & $72.0$ & - & $88.3$ \\
&Ramachandra \etal~\cite{Ramachandra-WACV-2020b} & $87.2$ & - & $94.0$ \\
&Sun \etal~\cite{Sun-ACMMM-2020} & $89.6$ & $74.7$ & - \\
&Tang \etal~\cite{Tang-PRL-2020} & $85.1$ & $73.0$ & $96.3$ \\
&Wang \etal~\cite{Wang-ACMMM-2020} & $87.0$ & $79.3$ & - \\
&Yu \etal~\cite{Yu-ACMMM-2020} & $89.6$ & $74.8$ & $97.3$ \\
&Zaheer \etal~\cite{Zaheer-CVPR-2020} & - & - & $98.1$ \\
\cline{2-5}
&Ours (object level) & $\color{blue}{91.9}$ & $\color{blue}{89.3}$ & $\color{red}{99.8}$ \\
&Ours (frame level)  & $86.9$ & $83.5$ & $92.4$ \\
&Ours (late fusion) & $\color{red}{92.8}$ & $\color{red}{90.2}$ & $\color{red}{99.8}$ \\
\hline
\end{tabular}
\end{center}
}
\vspace{-0.2cm}
\caption{Frame-level AUC scores (in \%) of the state-of-the-art methods \cite{Giorno-ECCV-2016,Dong-Access-2020,Doshi-CVPRW-2020a,Doshi-CVPRW-2020b,Gong-ICCV-2019,Hasan-CVPR-2016,Hinami-ICCV-2017,Ionescu-CVPR-2019,Ionescu-ICCV-2017,Ionescu-WACV-2019,Ji-IJCNN-2020,Kim-CVPR-2009,Lee-ICASSP-2018,Lee-TIP-2019,Liu-CVPR-2018,Liu-BMVC-2018,Lu-ICCV-2013,Lu-ECCV-2020,Luo-ICCV-2017,Mahadevan-CVPR-2010,Mehran-CVPR-2009,Nguyen-ICCV-2019,Park-CVPR-2020,Ramachandra-WACV-2020a,Ramachandra-WACV-2020b,Ravanbakhsh-WACV-2018,Ravanbakhsh-ICIP-2017,Smeureanu-ICIAP-2017,Sultani-CVPR-2018,Sun-ACMMM-2020,Tang-PRL-2020,Vu-AAAI-2019,Wang-ACMMM-2020,Wu-TNNLS-2019,Xu-BMVC-2015,Xu-CVIU-2017,Yu-ACMMM-2020,Zaheer-CVPR-2020,Zhang-PR-2016} versus our deep+wide architecture trained on four proxy tasks at the object level, at the frame level or based on late fusion. The top two results are shown in red and blue.\label{table:results}}
\vspace{-0.2cm}
\end{table}

\noindent
{\bf Parameter tuning.}
The first step of our framework is object detection based on YOLOv3 \cite{Redmon-arXiv-2018}. For Avenue and ShanghaiTech, we keep the detections with a confidence higher than $0.8$. Because UCSD Ped2 has a lower resolution, we set the detection confidence to $0.5$. We use the same confidence threshold during training and inference.

We use the first $85\%$ of the frames in each training video to train our models on the proxy tasks, keeping the last $15\%$ to validate the models on each proxy task. We fine-tune the parameters $t$ and $\lambda$ on our validation sets, before making the transition to anomaly detection. For $t$, we considered values in the set $\{1,2,3,4\}$. As we obtained optimal results with $t=3$, we use this value throughout all the anomaly detection experiments. Hence, an object-centric temporal sequence is a tensor of $7\times64\times64\times3$ components.
We fine-tune the parameter $\lambda$ controlling the importance of $\mathcal{L}_{T_4}$ in Equation~\eqref{eq_total_loss}, considering values in the set $\{0.1, 0.2, 0.5, 1.0\}$. We obtained optimal results with $\lambda=0.5$ on UCSD Ped2 and $\lambda=0.2$ on Avenue and ShanghaiTech, respectively. We therefore report anomaly detection results with these optimal settings.

Each neural network is trained for $30$ epochs using the Adam optimizer \cite{Kingma-ICLR-2014} with a learning rate of $10^{-3}$, keeping the default values for the other parameters of Adam. We trained the models using mini-batches of $256$ samples for the shallow+narrow architecture, $128$ samples for the deep+narrow and shallow+wide architectures and $64$ samples for the deep+wide architecture, being limited by our computational resources. For each model, we select the checkpoint with the lowest validation error on the proxy tasks to perform anomaly detection. 

\vspace*{-0.1cm}
\subsection{Anomaly Detection Results}
\vspace*{-0.1cm}

In Table \ref{table:results}, we present the comparative results of our object-level, frame-level and late fusion frameworks versus the state-of-the-art methods, reporting the frame-level AUC scores (whenever available) on the following three benchmark data sets: Avenue, ShanghaiTech and UCSD Ped2.

\begin{figure}[!t]
\begin{center}
\includegraphics[width=1.0\linewidth]{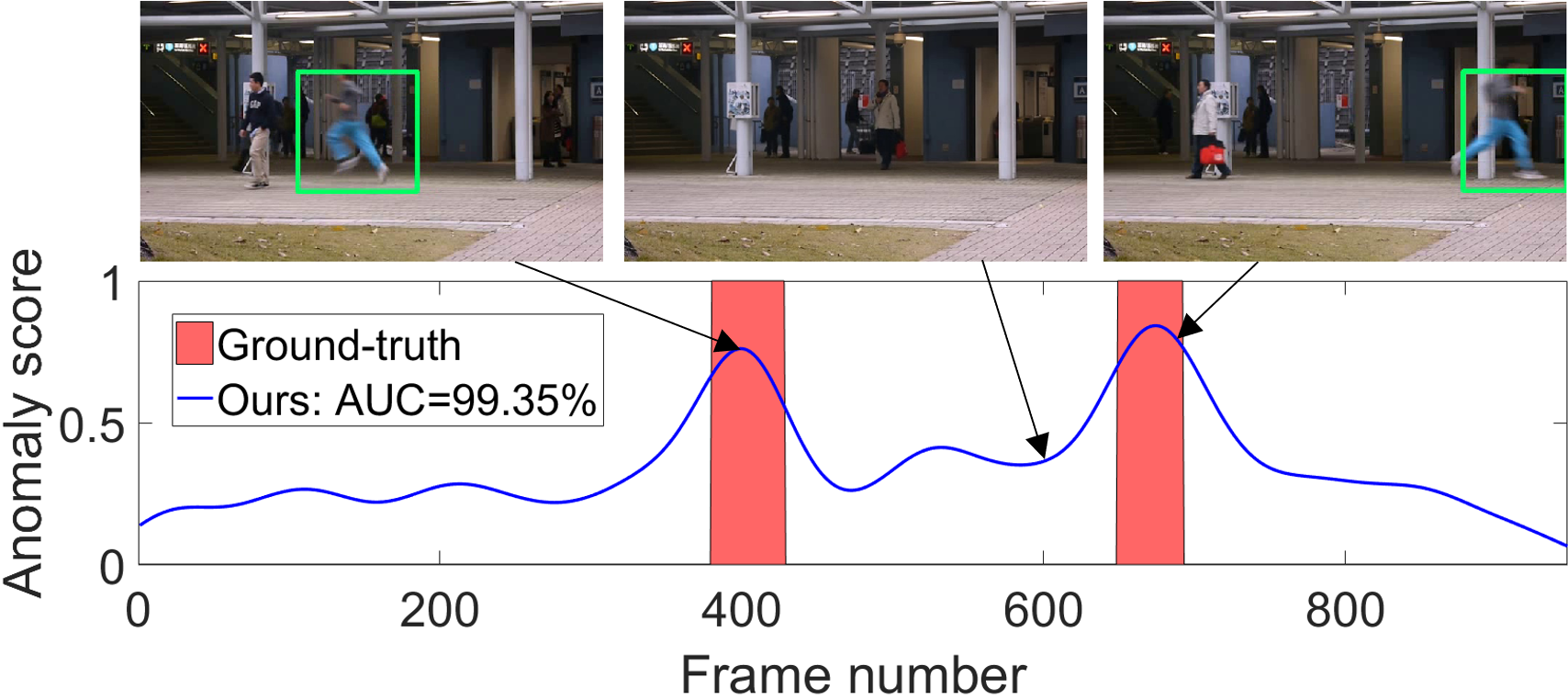}
\end{center}
\vspace{-0.4cm}
\caption{Frame-level scores and anomaly localization examples for test video 04 from Avenue. Best viewed in color.}
\label{fig:avenue}
\vspace{-0.1cm}
\end{figure}

\begin{figure}[!t]
\begin{center}
\includegraphics[width=1.0\linewidth]{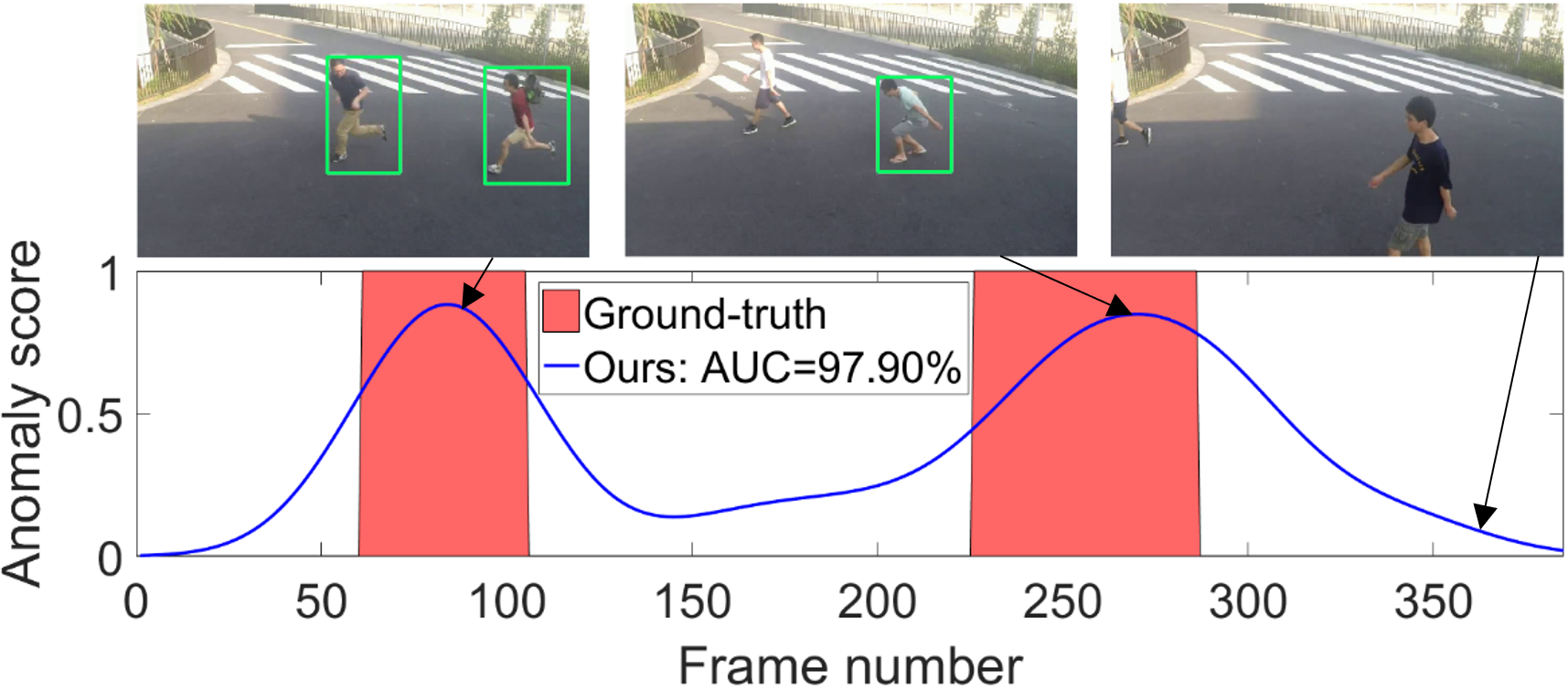}
\end{center}
\vspace{-0.35cm}
\caption{Frame-level scores and anomaly localization examples for test video 03{\_}0035 from ShanghaiTech. Best viewed in color.}
\label{fig:shanghai}
\vspace{-0.1cm}
\end{figure}

\begin{table}[t]
\small{
\begin{center}
\begin{tabular}{|l|c|c|c|}
\hline 
Method  & AUC & RBDC &  TBDC\\
\hline 
Ramachandra \etal~\cite{Ramachandra-WACV-2020a} & $88.3$ & $62.5$ & $80.5$ \\
Ramachandra \etal~\cite{Ramachandra-WACV-2020b}            & $94.0$ & $\color{red}{74.0}$ & $89.3$ \\
\hline
Ours (object level)                                                       & $\color{red}{99.8}$ & $72.8$ & $\color{red}{91.2}$ \\
\hline
\end{tabular}
\end{center}
}
\vspace*{-0.2cm}
\caption{Frame-level AUC, RBDC and TBDC scores (in \%) of two state-of-the-art methods \cite{Ramachandra-WACV-2020a,Ramachandra-WACV-2020b} versus our object-level framework. The best results are highlighted in red.\label{table:rbdc_tbdc}} 
\vspace*{-0.3cm}
\end{table}

\noindent
{\bf Results on Avenue.}
There are only two methods \cite{Ionescu-CVPR-2019, Lee-TIP-2019} that surpass the $90\%$ threshold on Avenue. Our framework applied at the object level obtains a frame-level AUC of $91.9\%$, surpassing the state-of-the-art method \cite{Ionescu-CVPR-2019} by $1.5\%$. When we apply our framework at the frame level, our performance drops considerably, but the method is still able to outperform some recent works \cite{Dong-Access-2020, Doshi-CVPRW-2020a, Ji-IJCNN-2020, Lu-ECCV-2020,Ramachandra-WACV-2020a, Tang-PRL-2020}. When we fuse the object-level anomaly scores with the frame-level anomaly scores, our performance improves, reaching a new state-of-the-art frame-level AUC of $92.8\%$.
In Figure~\ref{fig:avenue}, we illustrate a set of anomaly localization examples along with the frame-level anomaly scores for test video 04. We observe that our approach correlates well with the ground-truth frame-level annotations.

\begin{table*}[t]
\setlength\tabcolsep{5.0pt}
\small{
\begin{center}
\begin{tabular}{|c|l|l|c|c|c|c|c|c|c|c|c|c|}
\hline
Number & \multirow{3}{*}{3D CNN}      & \multirow{3}{*}{Level}       &  \multicolumn{5}{|c|}{Avenue} & \multicolumn{5}{|c|}{UCSD Ped2}    \\
\cline{4-13}                    
of & & &  \multicolumn{2}{|c|}{Accuracy}   &   \multicolumn{2}{|c|}{MAE} & \multirow{2}{*}{AUC} & \multicolumn{2}{|c|}{Accuracy}   &   \multicolumn{2}{|c|}{MAE} &  \multirow{2}{*}{AUC}  \\

\cline{4-7}  \cline{9-12} 
Tasks &           &       & Task 1 & Task 2 & Task 3 & Task 4 &  & Task 1 & Task 2 & Task 3 & Task 4 &  \\
\hline                 
1 & shallow+narrow & object & $84.8$ & - & - & - & $83.6$ & $98.1$ & -  & - & - & $89.4$\\
1 & shallow+narrow & object &  -     & $91.8$ & - & - & $83.4$     & -  & $99.3$ & - & - & $94.9$\\
1 & shallow+narrow & object &  -     & -  & $0.0001$ & - & $83.5$   & - & - & $0.0001$ & - & $97.1$ \\
1 & shallow+narrow & object &  -     & -  & -  & $0.0014$ & $73.7$ & - & - & - & $0.0014$& $97.1$ \\
\hline 
2 & shallow+narrow & object & $80.5$ & - & $0.0315$ & - & $87.7$ & $98.7$ & - & $0.0408$ & -  & $97.0$ \\
2 & deep+narrow    & object & $82.6$ & - & $0.0428$ & - & $83.7$ & $95.3$ & - & $0.0520$ & -  & $97.2$ \\
2 & shallow+wide   & object & $81.9$ & - & $0.0283$ & - & $83.7$ & $98.9$ & - & $0.0300$ & -  & $96.7$  \\
2 & deep+wide      & object & $82.4$ & - & $0.0383$ & - & $84.2$ & $98.5$ & - & $0.0554$ & - & $97.7$   \\
\hline
3 & shallow+narrow & object & $79.6$ & $89.6$ & $0.0350$ & - & $89.1$ & $98.0$ & $98.9$ & $0.0400$ & - & $97.5$ \\
3 & deep+narrow    & object & $89.9$ & $94.4$ & $0.0425$ & - & $91.6$ & $98.8$ & $99.7$ & $0.0501$ & - & $98.6$ \\ 
3 & shallow+wide   & object & $87.4$ & $93.3$ & $0.0305$ & - & $90.1$ & $98.8$ & $98.4$ & $0.0385$ & - & $97.5$ \\ 
3 & deep+wide      & object & $90.0$ & $95.2$ & $0.0410$ & - & $90.7$ & $98.9$ & $99.3$ & $0.0433$ & -  & $98.8$ \\ 
\hline
4 & shallow+narrow & object & $81.6$ & $92.2$ & $0.0337$ & $0.3898$ & $89.6$ & $98.7$ & $99.3$ & $0.0565$ & $0.3568$ & $99.1$ \\ 
4 & deep+narrow    & object & $89.6$ & $93.7$ & $0.0438$ & $0.3952$ & $91.5$ & $99.1$ & $98.4$ & $0.0499$ & $0.3807$ & $99.0$  \\ 
4 & shallow+wide   & object & $82.9$ & $91.0$ & $0.0313$ & $0.3767$ & $89.4$ & $98.8$ & $99.4$ & $0.0604$ & $0.3575$ & $97.8$  \\  
4 & deep+wide      & object & $92.2$ & $95.3$ & $0.0398$ & $0.3709$ & $\color{red}{91.9}$ & $99.0$ & $98.7$ & $0.0408$ & $0.3576$ & $\color{red}{99.8}$ \\
\hline
4 & deep+wide      & frame & $92.8$ & $96.1$ & $0.0199$ & $0.5608$ & $86.9$ & $99.9$ & $99.6$ & $0.0104$ & $0.4979$ & $92.4$  \\
\hline
\end{tabular}
\end{center}
}
\vspace*{-0.2cm}
\caption{Accuracy rates for Task 1 (arrow of time) and Task 2 (motion irregularity), mean absolute errors (MAE) for Task 3 (middle box prediction) and Task 4 (model distillation), and frame-level AUC scores (in \%) for anomaly detection obtained by adding one proxy task at a time. 
The best frame-level AUC scores are highlighted in red.\label{table:ablation}} 
\vspace*{-0.3cm}
\end{table*}

\noindent
{\bf Results on ShanghaiTech.}
On ShanghaiTech, our late fusion method outperforms all previous works, reaching a new state-of-the-art performance of $90.2\%$, surpassing the previous state-of-the-art method \cite{Ionescu-CVPR-2019} by a margin of $5.3\%$. Remarkably, we are the first to reach a frame-level AUC score of over $90\%$ on ShanghaiTech. Aside from \cite{Ionescu-CVPR-2019}, our method surpasses all other state-of-the-art approaches by a margin of at least $10.9\%$.
In Figure~\ref{fig:shanghai}, we present some anomaly localization examples along with the frame-level anomaly scores for test video 03{\_}0035. Our approach correlates well with the ground-truth annotations.

\noindent
{\bf Results on UCSD Ped2.}
UCSD Ped2 is one of the most popular video anomaly detection benchmarks, resulting in 23 works reporting frame-level AUC scores of over $90\%$. The current state-of-the-art method \cite{Vu-AAAI-2019} reports a frame-level AUC of $99.2\%$. Nevertheless, our method still manages to surpass all previous works, reaching a new state-of-the-art frame-level AUC of $99.8\%$ on UCSD Ped2.

Since RBDC and TBDC are part of a very recent evaluation protocol, there are only two methods \cite{Ramachandra-WACV-2020a,Ramachandra-WACV-2020b} that we can compare with in Table~\ref{table:rbdc_tbdc}. We outperform the first method \cite{Ramachandra-WACV-2020a} by significant margins in terms of all metrics. We also surpass the second method by $1.9\%$ in terms of TBDC and by $5.8\%$ in terms of frame-level AUC, our RBDC score being slightly lower. These results show that our approach can accurately localize anomalies.


\vspace*{-0.1cm}
\subsection{Ablation Study}
\vspace*{-0.1cm}

We perform an ablation study on Avenue and UCSD Ped2 to assess the benefit of including each proxy task in our joint multi-task framework. The corresponding results are presented in Table~\ref{table:ablation}. Along with the anomaly detection performance, we report the performance levels for each proxy task on our validation sets. Considering the individual tasks, we observe that the arrow of time produces the highest frame-level AUC ($83.6\%$) on Avenue, likely because anomalies are caused by unusual actions, \eg people running. The most suitable tasks for UCSD Ped2 seem to be middle bounding box prediction and knowledge distillation, probably because anomalies are caused by objects with unusual appearance, \eg bikes or cars.
We observe increasingly better anomaly detection results as we gradually add more proxy tasks in our joint optimization framework. 

While increasing the number of proxy tasks, we also aim to assess the effect of increasing the width and depth of our neural architecture. We observe performance improvements as we add more layers and filters to our 3D CNN, especially when we jointly optimize on three or four tasks. Hence, we conclude that it is beneficial to increase the learning capacity of the 3D CNN along with the number of proxy tasks. 


\vspace*{-0.1cm}
\section{Conclusion}
\label{sec_conclusion}
\vspace{-0.1cm}

In this work, we have proposed a novel anomaly detection method based on self-supervised and multi-task learning, presenting comprehensive results on three benchmarks: Avenue, ShanghaiTech and UCSD Ped2. To our knowledge, our method is the first and only to exceed the $90\%$ threshold on all three benchmarks. Additionally, we performed an ablation study showing the benefits of jointly learning multiple proxy tasks for anomaly detection in video. In future work, we will consider exploring additional proxy tasks to further boost the performance of our multi-task framework.

\vspace{-0.1cm}
\subsection*{Acknowledgments}
\vspace{-0.1cm}
The research leading to these results has received funding from the EEA Grants 2014-2021, under Project contract no.~EEA-RO-NO-2018-0496. This article has also benefited from the support of the Romanian Young Academy, which is funded by Stiftung Mercator and the Alexander von Humboldt Foundation for the period 2020-2022.

{\small
\bibliographystyle{ieee_fullname}
\bibliography{references}

\begin{thebibliography}{10}\itemsep=-1pt

\bibitem{Adam-PAMI-2008}
Amit Adam, Ehud Rivlin, Ilan Shimshoni, and Daviv Reinitz.
\newblock {Robust Real-Time Unusual Event Detection Using Multiple
  Fixed-Location Monitors}.
\newblock {\em IEEE Transactions on Pattern Analysis and Machine Intelligence},
  30(3):555--560, 2008.

\bibitem{Antic-ICCV-2011}
Borislav Antic and Bjorn Ommer.
\newblock Video parsing for abnormality detection.
\newblock In {\em Proceedings of ICCV}, pages 2415--2422, 2011.

\bibitem{Bergmann-CVPR-2020}
Paul Bergmann, Michael Fauser, David Sattlegger, and Carsten Steger.
\newblock {Uninformed Students: Student-Teacher Anomaly Detection With
  Discriminative Latent Embeddings}.
\newblock In {\em Proceedings of CVPR}, pages 4183--4192, 2020.

\bibitem{Cheng-CVPR-2015}
Kai-Wen Cheng, Yie-Tarng Chen, and Wen-Hsien Fang.
\newblock {Video anomaly detection and localization using hierarchical feature
  representation and Gaussian process regression}.
\newblock In {\em Proceedings of CVPR}, pages 2909--2917, 2015.

\bibitem{Cong-CVPR-2011}
Y. Cong, J. Yuan, and J. Liu.
\newblock Sparse reconstruction cost for abnormal event detection.
\newblock In {\em Proceedings of CVPR}, pages 3449--3456, 2011.

\bibitem{Cong-2013}
Yang Cong, Junsong Yuan, and Ji Liu.
\newblock Abnormal event detection in crowded scenes using sparse
  representation.
\newblock {\em Pattern Recognition}, 46:1851–1864, 07 2013.

\bibitem{Giorno-ECCV-2016}
Allison {Del Giorno}, {J. Andrew} Bagnell, and Martial Hebert.
\newblock {A Discriminative Framework for Anomaly Detection in Large Videos}.
\newblock In {\em Proceedings of ECCV}, pages 334--349, 2016.

\bibitem{Dong-Access-2020}
Fei Dong, Yu Zhang, and Xiushan Nie.
\newblock {Dual Discriminator Generative Adversarial Network for Video Anomaly
  Detection}.
\newblock {\em IEEE Access}, 8:88170--88176, 2020.

\bibitem{Doshi-CVPRW-2020a}
Keval Doshi and Yasin Yilmaz.
\newblock {Any-Shot Sequential Anomaly Detection in Surveillance Videos}.
\newblock In {\em Proceedings of CVPRW}, pages 934--935, 2020.

\bibitem{Doshi-CVPRW-2020b}
Keval Doshi and Yasin Yilmaz.
\newblock {Continual Learning for Anomaly Detection in Surveillance Videos}.
\newblock In {\em Proceedings of CVPRW}, pages 254--255, 2020.

\bibitem{Dutta-AAAI-2015}
Jayanta~K. Dutta and Bonny Banerjee.
\newblock {Online Detection of Abnormal Events Using Incremental Coding
  Length}.
\newblock In {\em Proceedings of AAAI}, pages 3755--3761, 2015.

\bibitem{Feng-NC-2017}
Yachuang Feng, Yuan Yuan, and Xiaoqiang Lu.
\newblock Learning deep event models for crowd anomaly detection.
\newblock {\em Neurocomputing}, 219:548--556, 2017.

\bibitem{Gong-ICCV-2019}
Dong Gong, Lingqiao Liu, Vuong Le, Budhaditya Saha, Moussa~Reda Mansour, Svetha
  Venkatesh, and Anton Van Den~Hengel.
\newblock {Memorizing Normality to Detect Anomaly: Memory-Augmented Deep
  Autoencoder for Unsupervised Anomaly Detection}.
\newblock In {\em Proceedings of ICCV}, pages 1705--1714, 2019.

\bibitem{Hasan-CVPR-2016}
Mahmudul Hasan, Jonghyun Choi, Jan Neumann, Amit~K. Roy{-}Chowdhury, and
  Larry~S. Davis.
\newblock Learning temporal regularity in video sequences.
\newblock In {\em Proceedings of CVPR}, pages 733--742, 2016.

\bibitem{He-CVPR-2016}
Kaiming He, Xiangyu Zhang, Shaoqing Ren, and Jian Sun.
\newblock {Deep Residual Learning for Image Recognition}.
\newblock In {\em Proceedings of CVPR}, pages 770--778, 2016.

\bibitem{Hinami-ICCV-2017}
Ryota Hinami, Tao Mei, and Shin'ichi Satoh.
\newblock {Joint Detection and Recounting of Abnormal Events by Learning Deep
  Generic Knowledge}.
\newblock In {\em Proceedings of ICCV}, pages 3639--3647, 2017.

\bibitem{Ionescu-CVPR-2019}
Radu~Tudor Ionescu, Fahad~Shahbaz Khan, Mariana-Iuliana Georgescu, and Ling
  Shao.
\newblock {Object-Centric Auto-Encoders and Dummy Anomalies for Abnormal Event
  Detection in Video}.
\newblock In {\em Proceedings of CVPR}, pages 7842--7851, 2019.

\bibitem{Ionescu-ICCV-2017}
Radu~Tudor Ionescu, Sorina Smeureanu, Bogdan Alexe, and Marius Popescu.
\newblock Unmasking the abnormal events in video.
\newblock In {\em Proceedings of ICCV}, pages 2895--2903, 2017.

\bibitem{Ionescu-WACV-2019}
Radu~Tudor Ionescu, Sorina Smeureanu, Marius Popescu, and Bogdan Alexe.
\newblock {Detecting abnormal events in video using Narrowed Normality
  Clusters}.
\newblock In {\em Proceedings of WACV}, pages 1951--1960, 2019.

\bibitem{Ji-IJCNN-2020}
Xiangli Ji, Bairong Li, and Yuesheng Zhu.
\newblock {TAM-Net: Temporal Enhanced Appearance-to-Motion Generative Network
  for Video Anomaly Detection}.
\newblock In {\em Proceedings of IJCNN}, pages 1--8, 2020.

\bibitem{Kim-CVPR-2009}
Jaechul Kim and Kristen Grauman.
\newblock {Observe locally, infer globally: A space-time MRF for detecting
  abnormal activities with incremental updates}.
\newblock In {\em Proceedings of CVPR}, pages 2921--2928, 2009.

\bibitem{Kingma-ICLR-2014}
Diederik~P. Kingma and Jimmy Ba.
\newblock Adam: A method for stochastic optimization.
\newblock In {\em Proceedings of ICLR}, 2015.

\bibitem{Lee-ICASSP-2018}
Sangmin Lee, Hak~Gu Kim, and Yong~Man Ro.
\newblock {STAN: Spatio-temporal adversarial networks for abnormal event
  detection}.
\newblock In {\em Proceedings of ICASSP}, pages 1323--1327, 2018.

\bibitem{Lee-TIP-2019}
Sangmin Lee, Hak~Gu Kim, and Yong~Man Ro.
\newblock {BMAN: Bidirectional Multi-Scale Aggregation Networks for Abnormal
  Event Detection}.
\newblock {\em IEEE Transactions on Image Processing}, 29:2395--2408, 2019.

\bibitem{Li-PAMI-2014}
Weixin Li, Vijay Mahadevan, and Nuno Vasconcelos.
\newblock Anomaly detection and localization in crowded scenes.
\newblock {\em IEEE Transactions on Pattern Analysis and Machine Intelligence},
  36(1):18--32, 2014.

\bibitem{Lin-ECCV-2014}
Tsung-Yi Lin, Michael Maire, Serge Belongie, James Hays, Pietro Perona, Deva
  Ramanan, Piotr Doll{\'a}r, and C~Lawrence Zitnick.
\newblock {Microsoft COCO: Common Objects in Context}.
\newblock In {\em Proceedings of ECCV}, pages 740--755, 2014.

\bibitem{Liu-CVPR-2018}
Wen Liu, Weixin Luo, Dongze Lian, and Shenghua Gao.
\newblock {Future Frame Prediction for Anomaly Detection -- A New Baseline}.
\newblock In {\em Proceedings of CVPR}, pages 6536--6545, 2018.

\bibitem{Liu-BMVC-2018}
Yusha Liu, Chun-Liang Li, and Barnaba\'{a}s P\'{o}czos.
\newblock {Classifier Two-Sample Test for Video Anomaly Detections}.
\newblock In {\em Proceedings of BMVC}, 2018.

\bibitem{Lu-ICCV-2013}
C. Lu, J. Shi, and J. Jia.
\newblock {Abnormal Event Detection at 150 FPS in MATLAB}.
\newblock In {\em Proceedings of ICCV}, pages 2720--2727, 2013.

\bibitem{Lu-ECCV-2020}
Yiwei Lu, Frank Yu, Mahesh Kumar, Krishna Reddy, and Yang Wang.
\newblock {Few-Shot Scene-Adaptive Anomaly Detection}.
\newblock In {\em Proceedings of ECCV}, pages 125--141, 2020.

\bibitem{Luo-ICCV-2017}
Weixin Luo, Wen Liu, and Shenghua Gao.
\newblock {A Revisit of Sparse Coding Based Anomaly Detection in Stacked RNN
  Framework}.
\newblock In {\em Proceedings of ICCV}, pages 341--349, 2017.

\bibitem{Mahadevan-CVPR-2010}
Vijay Mahadevan, Wei-Xin LI, Viral Bhalodia, and Nuno Vasconcelos.
\newblock {Anomaly Detection in Crowded Scenes}.
\newblock In {\em Proceedings of CVPR}, pages 1975--1981, 2010.

\bibitem{Mehran-CVPR-2009}
Ramin Mehran, Alexis Oyama, and Mubarak Shah.
\newblock Abnormal crowd behavior detection using social force model.
\newblock In {\em Proceedings of CVPR}, pages 935--942, 2009.

\bibitem{Nguyen-ICCV-2019}
Trong-Nguyen Nguyen and Jean Meunier.
\newblock Anomaly detection in video sequence with appearance-motion
  correspondence.
\newblock In {\em Proceedings of ICCV}, pages 1273--1283, 2019.

\bibitem{Pang-CVPR-2020}
Guansong Pang, Cheng Yan, Chunhua Shen, Anton van~den Hengel, and Xiao Bai.
\newblock {Self-trained Deep Ordinal Regression for End-to-End Video Anomaly
  Detection}.
\newblock In {\em Proceedings of CVPR}, pages 12173--12182, 2020.

\bibitem{Park-CVPR-2020}
Hyunjong Park, Jongyoun Noh, and Bumsub Ham.
\newblock {Learning Memory-guided Normality for Anomaly Detection}.
\newblock In {\em Proceedings of CVPR}, pages 14372--14381, 2020.

\bibitem{Ramachandra-WACV-2020a}
Bharathkumar Ramachandra and Michael Jones.
\newblock {Street Scene: A new dataset and evaluation protocol for video
  anomaly detection}.
\newblock In {\em Proceedings of WACV}, pages 2569--2578, 2020.

\bibitem{Ramachandra-WACV-2020b}
Bharathkumar Ramachandra, Michael Jones, and Ranga Vatsavai.
\newblock {Learning a distance function with a Siamese network to localize
  anomalies in videos}.
\newblock In {\em Proceedings of WACV}, pages 2598--2607, 2020.

\bibitem{Ramachandra-ArXiv-2020}
Bharathkumar Ramachandra, Michael~J. Jones, and Ranga~Raju Vatsavai.
\newblock {A Survey of Single-Scene Video Anomaly Detection}.
\newblock {\em arXiv preprint arXiv:2004.05993}, 2020.

\bibitem{Ravanbakhsh-WACV-2018}
Mahdyar Ravanbakhsh, Moin Nabi, Hossein Mousavi, Enver Sangineto, and Nicu
  Sebe.
\newblock {Plug-and-Play CNN for Crowd Motion Analysis: An Application in
  Abnormal Event Detection}.
\newblock In {\em Proceedings of WACV}, pages 1689--1698, 2018.

\bibitem{Ravanbakhsh-ICIP-2017}
Mahdyar Ravanbakhsh, Moin Nabi, Enver Sangineto, Lucio Marcenaro, Carlo
  Regazzoni, and Nicu Sebe.
\newblock {Abnormal Event Detection in Videos using Generative Adversarial
  Nets}.
\newblock In {\em Proceedings of ICIP}, pages 1577--1581, 2017.

\bibitem{Redmon-arXiv-2018}
Joseph Redmon and Ali Farhadi.
\newblock {YOLOv3: An incremental improvement}.
\newblock {\em arXiv preprint arXiv:1804.02767}, 2018.

\bibitem{Russakovsky2015}
O. Russakovsky, J. Deng, H. Su, J. Krause, S. Satheesh, S. Ma, Z. Huang,
  Karpathy A., A. Khosla, M. Bernstein, A.~C. Berg, and L. Fei-Fei.
\newblock {ImageNet Large Scale Visual Recognition Challenge}.
\newblock {\em International Journal of Computer Vision}, 115(3):211--252,
  2015.

\bibitem{Sabokrou-IP-2017}
Mohammad Sabokrou, Mohsen Fayyaz, Mahmood Fathy, and Reinhard Klette.
\newblock {Deep-Cascade: Cascading 3D Deep Neural Networks for Fast Anomaly
  Detection and Localization in Crowded Scenes}.
\newblock {\em IEEE Transactions on Image Processing}, 26(4):1992--2004, 2017.

\bibitem{Sabokrou-CVIU-2018}
Mohammad Sabokrou, Mohsen Fayyaz, Mahmood Fathy, Zahra Moayed, and Reinhard
  Klette.
\newblock Deep-anomaly: Fully convolutional neural network for fast anomaly
  detection in crowded scenes.
\newblock {\em Computer Vision and Image Understanding}, 172:88--97, 2018.

\bibitem{Saligrama-CVPR-2012}
Venkatesh Saligrama and Zhu Chen.
\newblock Video anomaly detection based on local statistical aggregates.
\newblock In {\em Proceedings of CVPR}, pages 2112--2119, 2012.

\bibitem{Smeureanu-ICIAP-2017}
Sorina Smeureanu, Radu~Tudor Ionescu, Marius Popescu, and Bogdan Alexe.
\newblock {Deep Appearance Features for Abnormal Behavior Detection in Video}.
\newblock In {\em Proceedings of ICIAP}, volume 10485, pages 779--789, 2017.

\bibitem{Sultani-CVPR-2018}
Waqas Sultani, Chen Chen, and Mubarak Shah.
\newblock {Real-World Anomaly Detection in Surveillance Videos}.
\newblock In {\em Proceedings of CVPR}, pages 6479--6488, 2018.

\bibitem{Sun-ACMMM-2020}
Che Sun, Yunde Jia, Yao Hu, and Yuwei Wu.
\newblock {Scene-Aware Context Reasoning for Unsupervised Abnormal Event
  Detection in Videos}.
\newblock In {\em Proceedings of ACMMM}, pages 184--192, 2020.

\bibitem{Sun-PR-2017}
Qianru Sun, Hong Liu, and Tatsuya Harada.
\newblock Online growing neural gas for anomaly detection in changing
  surveillance scenes.
\newblock {\em Pattern Recognition}, 64(C):187--201, Apr. 2017.

\bibitem{Tang-PRL-2020}
Yao Tang, Lin Zhao, Shanshan Zhang, Chen Gong, Guangyu Li, and Jian Yang.
\newblock Integrating prediction and reconstruction for anomaly detection.
\newblock {\em Pattern Recognition Letters}, 129:123--130, 2020.

\bibitem{Tran-BMVC-2017}
Hanh~T.M. Tran and David Hogg.
\newblock {Anomaly Detection using a Convolutional Winner-Take-All
  Autoencoder}.
\newblock In {\em Proceedings of BMVC}, 2017.

\bibitem{Vu-AAAI-2019}
Hung Vu, Tu~Dinh Nguyen, Trung Le, Wei Luo, and Dinh Phung.
\newblock {Robust Anomaly Detection in Videos Using Multilevel
  Representations}.
\newblock In {\em Proceedings of AAAI}, volume~33, pages 5216--5223, 2019.

\bibitem{Wang-ICIP-2018}
L. {Wang}, F. {Zhou}, Z. {Li}, W. {Zuo}, and H. {Tan}.
\newblock {Abnormal Event Detection in Videos Using Hybrid Spatio-Temporal
  Autoencoder}.
\newblock In {\em Proceedings of ICIP}, pages 2276--2280, 2018.

\bibitem{Wang-ACMMM-2020}
Ziming Wang, Yuexian Zou, and Zeming Zhang.
\newblock {Cluster Attention Contrast for Video Anomaly Detection}.
\newblock In {\em Proceedings of ACMMM}, pages 2463--2471, 2020.

\bibitem{Wei-CVPR-2018}
Donglai Wei, Joseph~J. Lim, Andrew Zisserman, and William~T. Freeman.
\newblock {Learning and Using the Arrow of Time}.
\newblock In {\em Proceedings of CVPR}, pages 8052--8060, 2018.

\bibitem{Wu-TNNLS-2019}
Peng Wu, Jing Liu, and Fang Shen.
\newblock {A Deep One-Class Neural Network for Anomalous Event Detection in
  Complex Scenes}.
\newblock {\em IEEE Transactions on Neural Networks and Learning Systems},
  31(7):2609--2622, 2019.

\bibitem{Wu-CVPR-2010}
Shandong Wu, Brian~E. Moore, and Mubarak Shah.
\newblock {Chaotic Invariants of Lagrangian Particle Trajectories for Anomaly
  Detection in Crowded Scenes}.
\newblock In {\em Proceedings of CVPR}, pages 2054--2060, 2010.

\bibitem{Xu-BMVC-2015}
Dan Xu, Elisa Ricci, Yan Yan, Jingkuan Song, and Nicu Sebe.
\newblock {Learning Deep Representations of Appearance and Motion for Anomalous
  Event Detection}.
\newblock In {\em Proceedings of BMVC}, pages 8.1--8.12, 2015.

\bibitem{Xu-CVIU-2017}
Dan Xu, Yan Yan, Elisa Ricci, and Nicu Sebe.
\newblock {Detecting Anomalous Events in Videos by Learning Deep
  Representations of Appearance and Motion}.
\newblock {\em Computer Vision and Image Understanding}, 156:117--127, 2017.

\bibitem{Yu-ACMMM-2020}
Guang Yu, Siqi Wang, Zhiping Cai, En Zhu, Chuanfu Xu, Jianping Yin, and Marius
  Kloft.
\newblock {Cloze Test Helps: Effective Video Anomaly Detection via Learning to
  Complete Video Events}.
\newblock In {\em Proceedings of ACMMM}, pages 583--591, 2020.

\bibitem{Zaheer-CVPR-2020}
Muhammad~Zaigham Zaheer, Jin-ha Lee, Marcella Astrid, and Seung-Ik Lee.
\newblock {Old is Gold: Redefining the Adversarially Learned One-Class
  Classifier Training Paradigm}.
\newblock In {\em Proceedings of CVPR}, pages 14183--14193, 2020.

\bibitem{Zhang-PR-2020}
Xinfeng Zhang, Su Yang, Jiulong Zhang, and Weishan Zhang.
\newblock {Video Anomaly Detection and Localization using Motion-field Shape
  Description and Homogeneity Testing}.
\newblock {\em Pattern Recognition}, page 107394, 2020.

\bibitem{Zhang-PR-2016}
Ying Zhang, Huchuan Lu, Lihe Zhang, Xiang Ruan, and Shun Sakai.
\newblock Video anomaly detection based on locality sensitive hashing filters.
\newblock {\em Pattern Recognition}, 59:302--311, 2016.

\bibitem{Zhong-CVPR-2019}
Jia-Xing Zhong, Nannan Li, Weijie Kong, Shan Liu, Thomas~H. Li, and Ge Li.
\newblock {Graph Convolutional Label Noise Cleaner: Train a Plug-And-Play
  Action Classifier for Anomaly Detection}.
\newblock In {\em Proceedings of CVPR}, pages 1237--1246, 2019.

\end{thebibliography}
}

\section{Supplementary}

\begin{figure*}[!th]
\begin{center}
\includegraphics[width=1.0\linewidth]{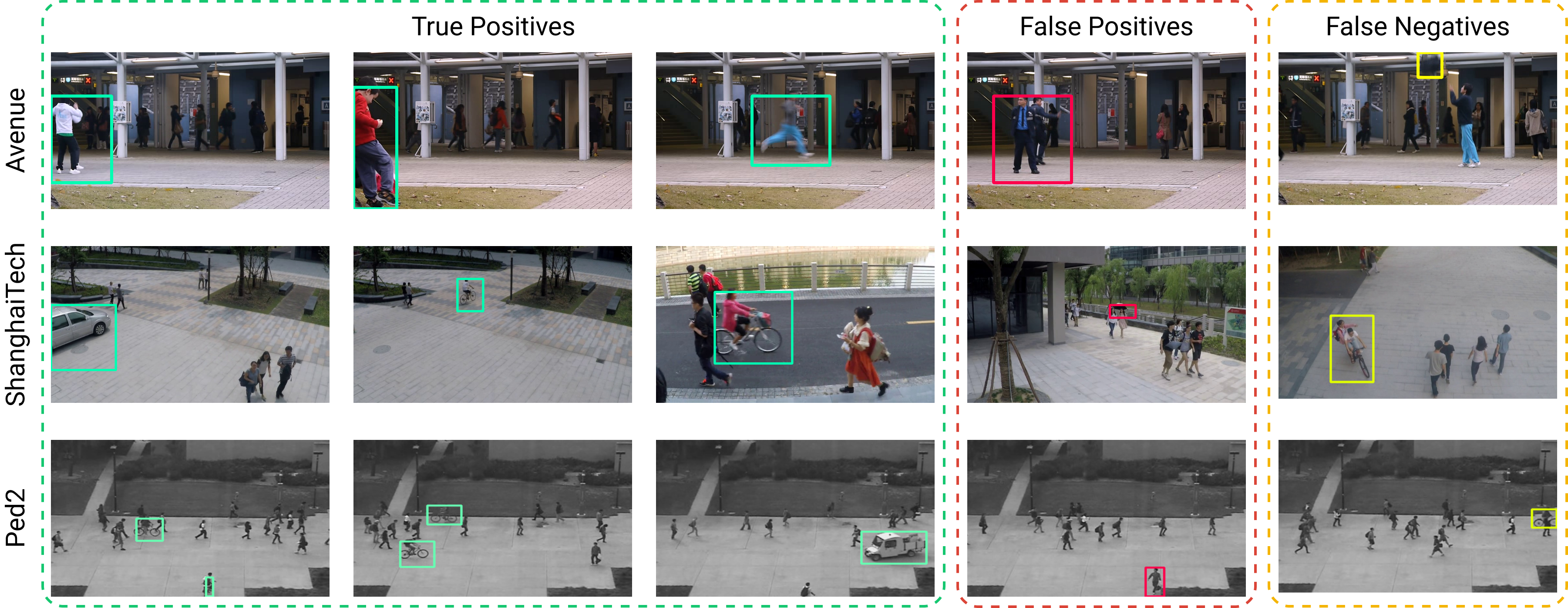}
\end{center}
\vspace{-0.4cm}
\caption{True positive, false positive and false negative examples from Avenue (top row), ShanghaiTech (second row) and UCSD Ped2 (bottom row). Best viewed in color.}
\label{fig:tp}
\vspace{-0.1cm}
\end{figure*}

In the supplementary, we include additional examples of frame-level scores predicted by our object-centric framework. Along with the frame-level scores, we also show anomaly localization examples in specific frames. Besides showing correct detections, we also include a set of false positive and false negative examples. Moreover, the supplementary provides details about the running time and a discussion about the reliance on object detectors and the chosen proxy tasks.

\begin{figure*}[!t]
\begin{center}
\includegraphics[width=0.9\linewidth]{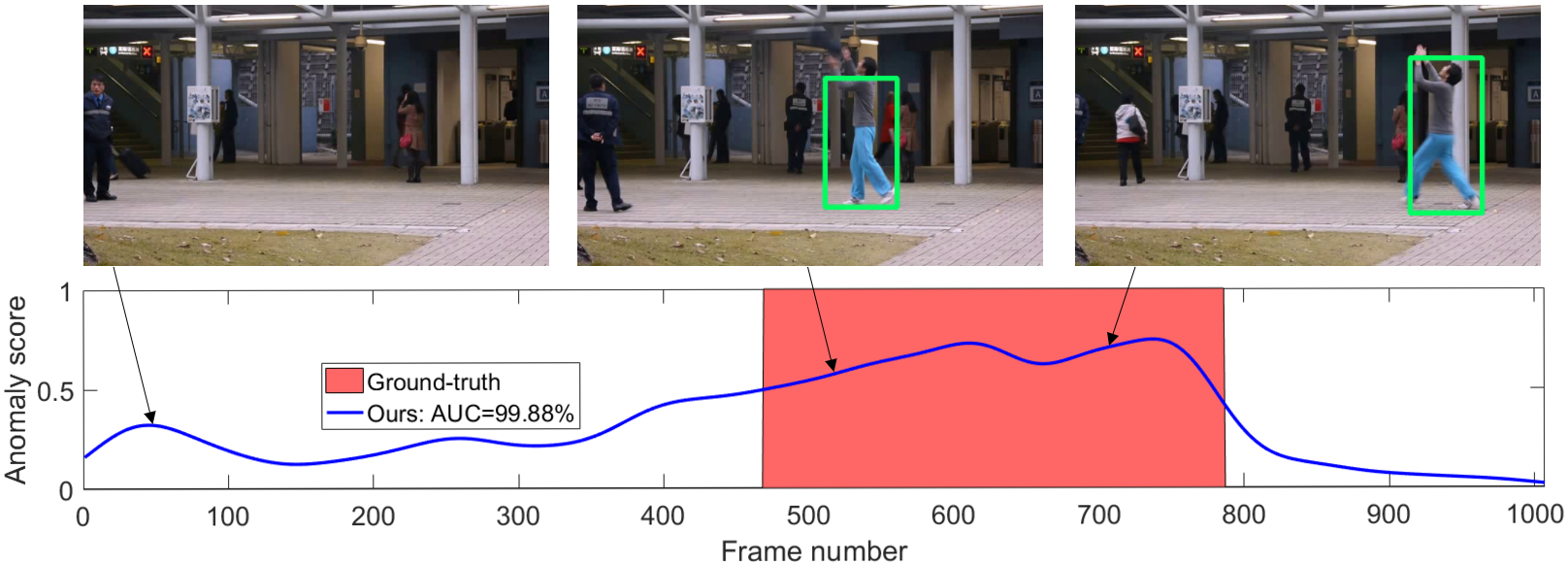}
\end{center}
\vspace{-0.4cm}
\caption{Frame-level scores and anomaly localization examples for test video 05 from Avenue. Best viewed in color.}
\label{fig:avenue_05}
\vspace{-0.1cm}
\end{figure*} 

\begin{figure*}[!t]
\begin{center}
\includegraphics[width=0.95\linewidth]{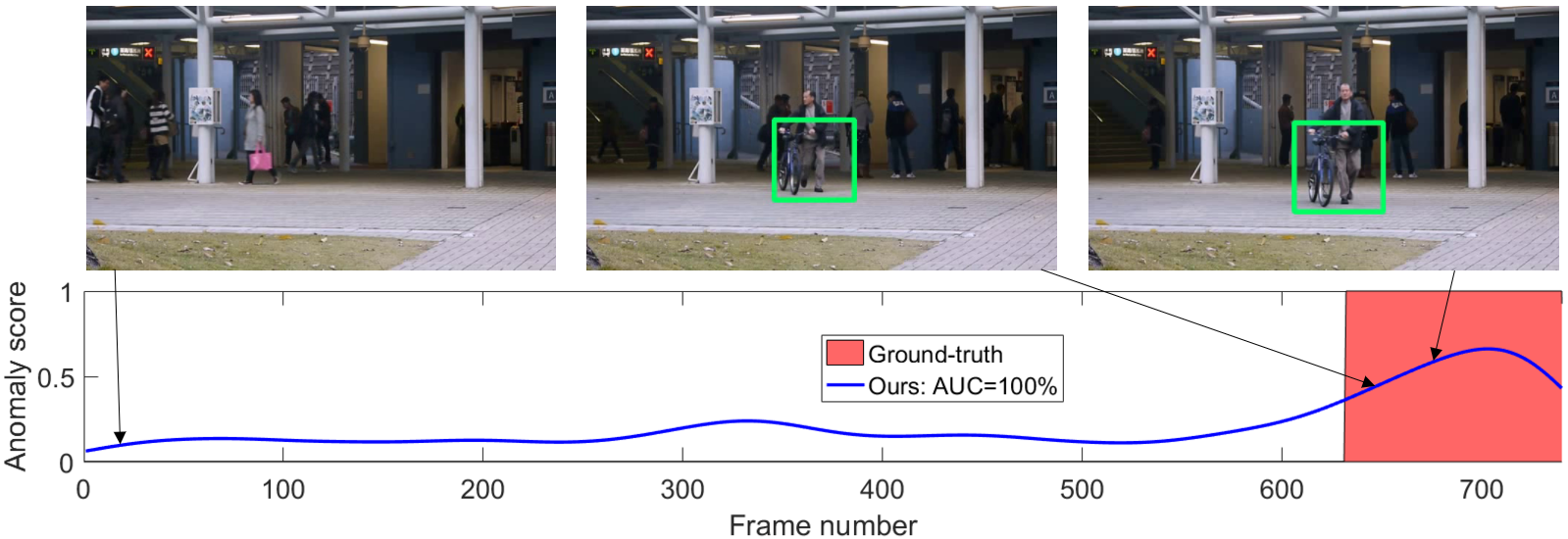}
\end{center}
\vspace{-0.4cm}
\caption{Frame-level scores and anomaly localization examples for test video 16 from Avenue. Best viewed in color.}
\label{fig:avenue_16}
\vspace{-0.1cm}
\end{figure*}

\begin{figure*}[!t]
\begin{center}
\includegraphics[width=0.95\linewidth]{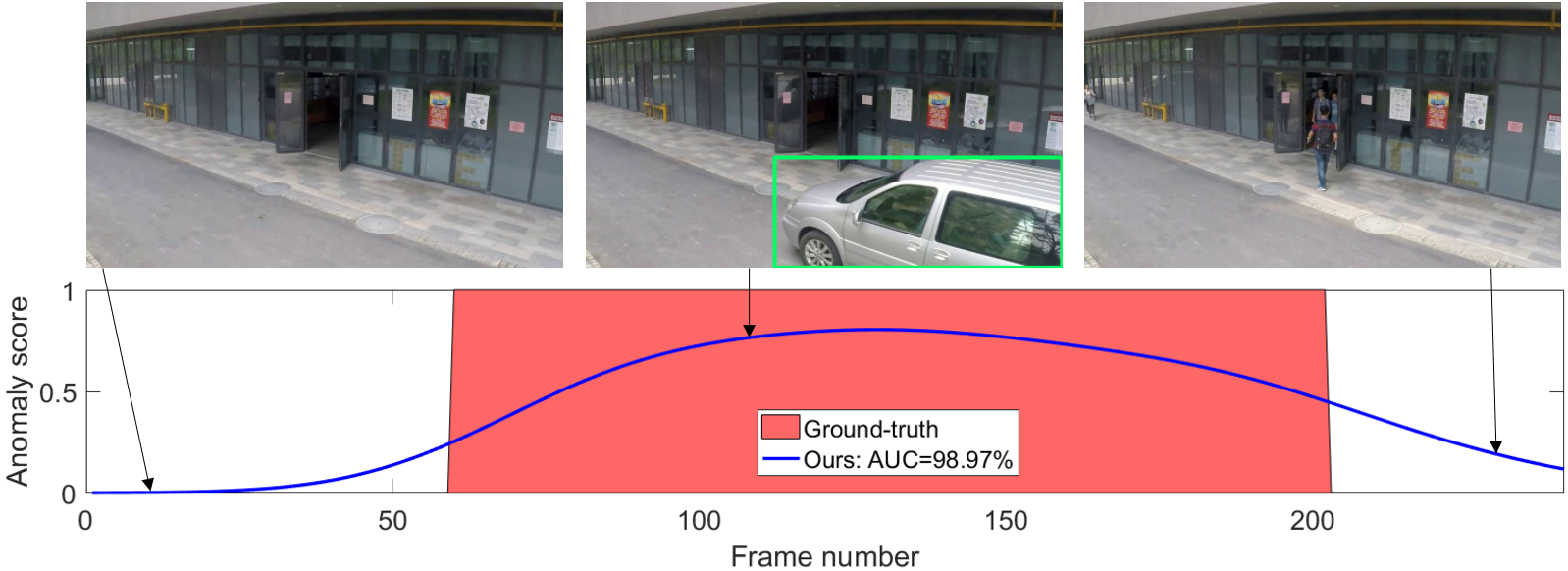}
\end{center}
\vspace{-0.4cm}
\caption{Frame-level scores and anomaly localization examples for test video 06{\_}0144 from ShanghaiTech. Best viewed in color.}
\label{fig:shanghai_06_0144}
\vspace{-0.1cm}
\end{figure*}

\begin{figure*}[!t]
\begin{center}
\includegraphics[width=0.95\linewidth]{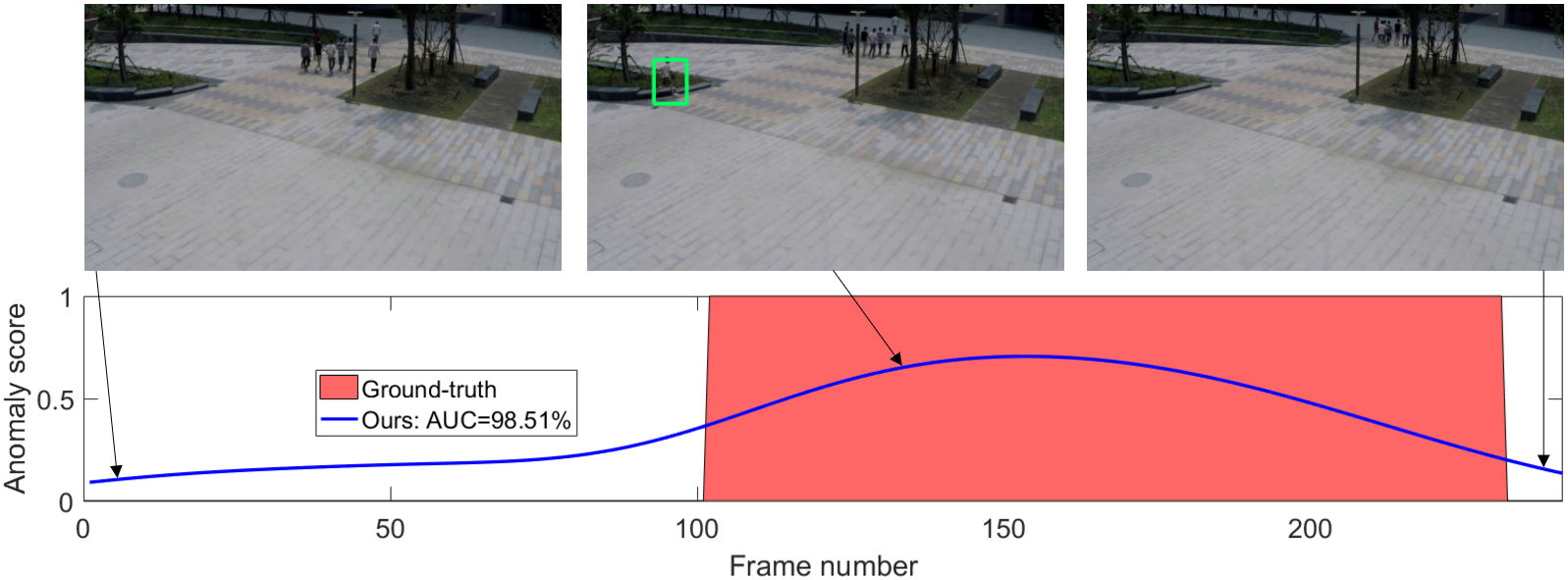}
\end{center}
\vspace{-0.4cm}
\caption{Frame-level scores and anomaly localization examples for test video 12{\_}0149 from ShanghaiTech. Best viewed in color.}
\label{fig:shanghai_12_0149}
\vspace{-0.1cm}
\end{figure*}

\begin{figure*}[!t]
\begin{center}
\includegraphics[width=0.95\linewidth]{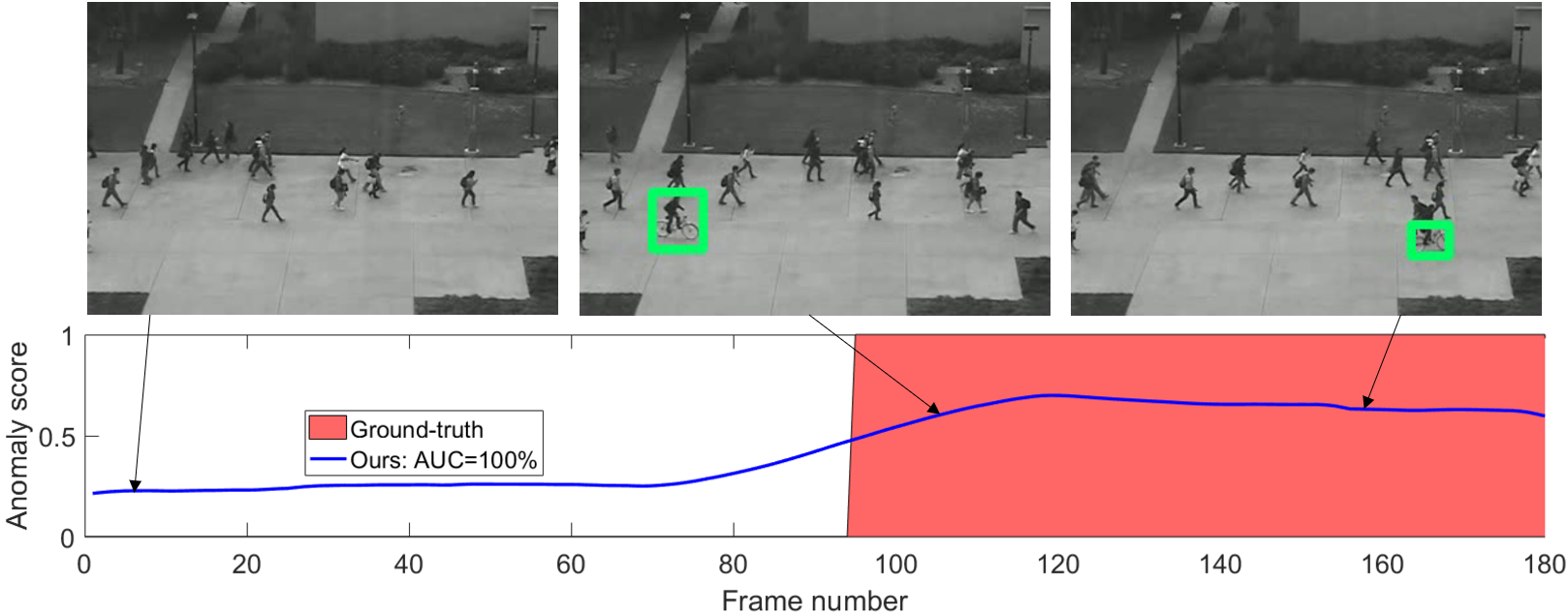}
\end{center}
\vspace{-0.4cm}
\caption{Frame-level scores and anomaly localization examples for test video 02 from UCSD Ped2. Best viewed in color.}
\label{fig:ped_02}
\vspace{-0.1cm}
\end{figure*}

\begin{figure*}[!t]
\begin{center}
\includegraphics[width=0.95\linewidth]{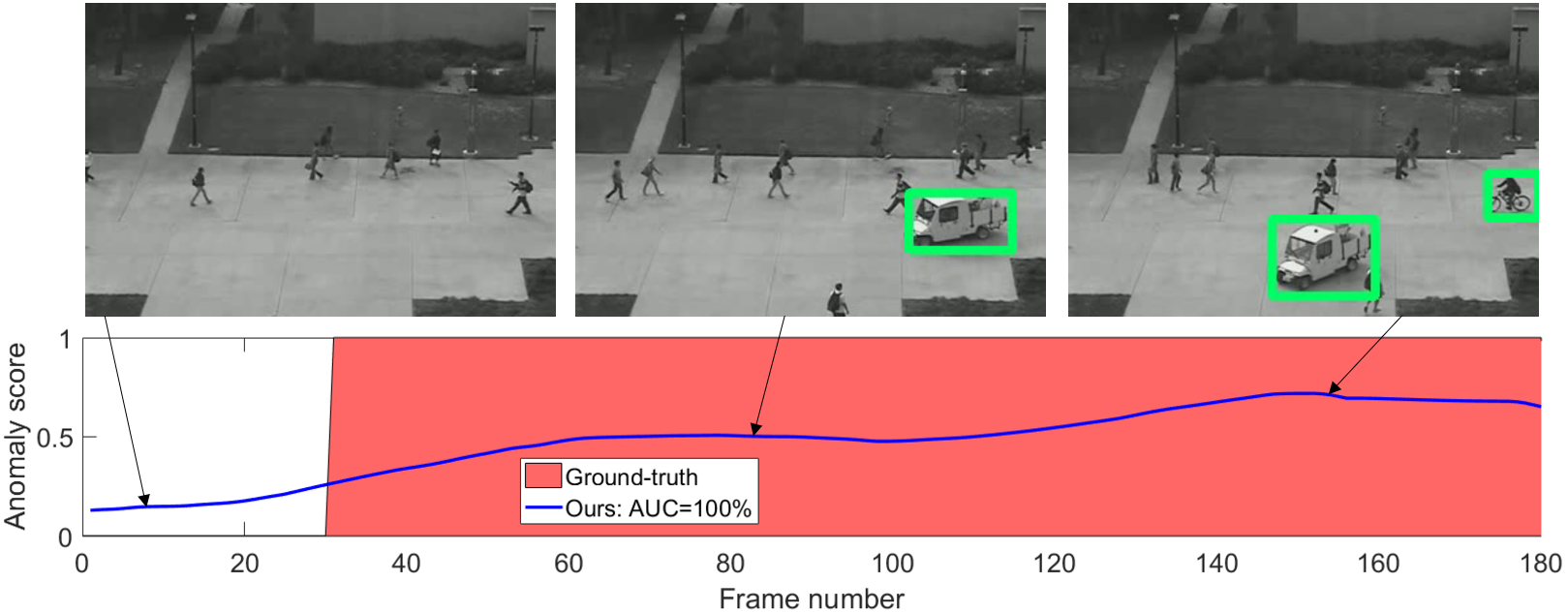}
\end{center}
\vspace{-0.4cm}
\caption{Frame-level scores and anomaly localization examples for test video 04 from UCSD Ped2. Best viewed in color.}
\label{fig:ped_04}
\vspace{-0.1cm}
\end{figure*}

\begin{figure*}[!t]
\begin{center}
\includegraphics[width=0.95\linewidth]{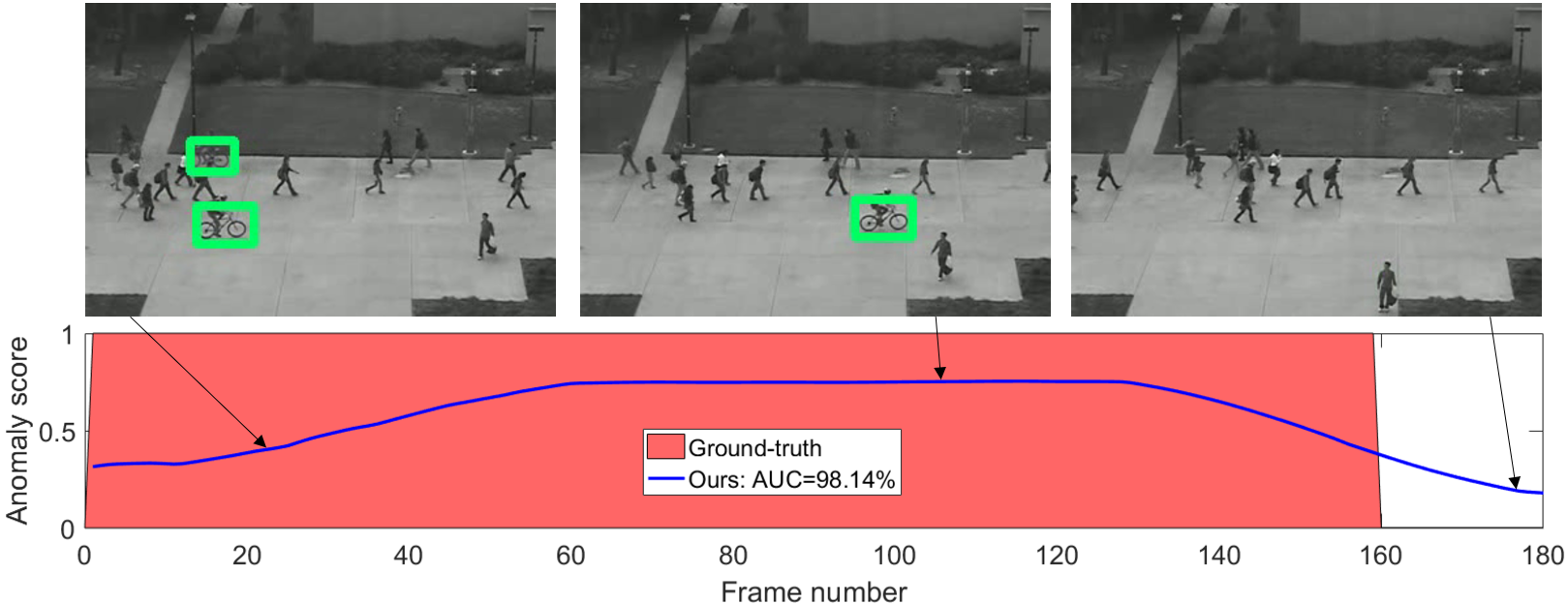}
\end{center}
\vspace{-0.4cm}
\caption{Frame-level scores and anomaly localization examples for test video 06 from UCSD Ped2. Best viewed in color.}
\label{fig:ped_06}
\vspace{-0.1cm}
\end{figure*}


\subsection{Qualitative Results}

The supplementary results are structured as follows. Figure \ref{fig:tp} illustrates a set of true positive, false positive and false negative examples extracted from our runs on the benchmark data sets. Figures \ref{fig:avenue_05} and \ref{fig:avenue_16} showcase the overlap between our frame-level anomaly predictions and the ground-truth labels for two videos from Avenue. Similarly, Figures \ref{fig:shanghai_06_0144} and \ref{fig:shanghai_12_0149} illustrate the overlap between our frame-level anomaly predictions and the ground-truth labels for two ShanghaiTech videos. Finally, Figures \ref{fig:ped_02}, \ref{fig:ped_04} and \ref{fig:ped_06} showcase our frame-level performance for three UCSD Ped2 videos.

\paragraph{Avenue.} Our framework reaches a state-of-the-art frame-level AUC performance of $92.8\%$ on the Avenue data set, being able to detect anomalies such as: $(i)$ the two, mostly overlapped, individuals dressed in white preforming a dance on one side of the scene, $(ii)$ the child dressed in red that was moving very close to the camera and $(iii)$ the man running on the main alley, all shown in Figure \ref{fig:tp} (top row). Aside from these true positive detections, we present a false positive example of two people that act strangely. 
In this specific instance, the security agent that took a stance in front of the main alley was wrongly labeled as anomalous, probably because this behavior is not observed during training. Finally, due to the detection failure of the object detector, our framework is not able to label the backpack thrown in the air as an anomaly, generating the false negative illustrated in Figure \ref{fig:tp} (top row). This deficiency is compensated by recognizing that the gesture of throwing a backpack into the air performed by the human is indeed anomalous. Figure \ref{fig:avenue_05} illustrates how our framework is able to capture the gesture of throwing, labeling the individual as anomalous. Our framework reaches an almost perfect frame-level AUC performance of $99.88\%$ on the fifth test video from the Avenue data set. Additionally, Figure \ref{fig:avenue_16} showcases how our framework is able to detect other object-related anomalies. In this instance, our anomaly score starts to increase as the bike appears in the scene. Our method reports it as a clear anomalous occurrence as it becomes fully visible and moves towards the camera.

\paragraph{ShanghaiTech.} On ShanghaiTech, our framework is able to correctly identify most vehicle-related anomalies. As show in Figure \ref{fig:tp} (second row), objects such as cars and bicycles are regularly labeled as anomalies. However, in the specific scenario presented as false negative in Figure \ref{fig:tp} (second row), a bicycle that was used by two individuals simultaneously managed to pass as a normal event. Aside from vehicles, our framework also labels strange (meaning not previously seen) objects as anomalies when encountered. Accordingly, in the false positive example, the umbrella was detected and labeled as anomalous. Figures \ref{fig:shanghai_06_0144} and \ref{fig:shanghai_12_0149} showcase our anomaly score predictions together with the frame-level ground-truth labels for test videos \textit{06\_0144} and \textit{12\_0149} from ShanghaiTech, respectively. In the first instance, our method correctly identifies the car as an anomaly, reaching a frame-level AUC of $98.97\%$, while in the second instance, our framework accurately identifies the individual running behind the group as abnormal, reaching a frame-level AUC of $98.51\%$. 

\paragraph{UCSD Ped2.} On UCSD Ped2, our method reaches a frame-level AUC of $99.8\%$, accurately and almost perfectly capturing all anomalous events such as people riding bicycles among the crowd or vehicles making an appearance in the pedestrian area. Objects are missed only in very few particular frames, such as when the bike did not completely enter the scene (being truncated), shown as the false negative example from UCSD Ped2 in Figure \ref{fig:tp} (bottom row). In addition, the individual featured as the false positive leaving the alley through the camera-facing exit is also wrongly labeled as an anomaly. Figures \ref{fig:ped_02} and \ref{fig:ped_04} showcase the general performance of our method on the UCSD Ped2 data set, reaching perfect frame-level AUC scores.

\subsection{Running Time}

Our lightweight model infers the anomaly score of a single object in $6$ milliseconds (ms). The YOLOv3 model takes $26$ ms per frame to detect the objects. Reassembling the anomaly map from the object-level anomaly scores takes less than $1$ ms. With all components in place, our framework runs at $23$ FPS with an average of $5$ objects per frame. The reported time includes only the object-level inference, which is the most heavy part (due to the object detector). When we add the frame-level inference, the speed decreases by a small margin, from 23 FPS to 21 FPS. The FPS rates are measured on a single GeForce GTX 1080Ti GPU with 11GB of VRAM.

\subsection{Discussion}

\paragraph{Dependence on object detector.}
We note that object-centric methods are influenced by the quality of object detectors. For example, on Avenue, we observed that our object-centric method does not detect papers (\emph{paper} is not in the COCO set of classes) or backpacks thrown in the air (\emph{backpack} is in the COCO set of classes, but the detector fails due to motion blur). Despite not explicitly detecting papers or backpacks, the detector detects the person throwing these objects and our framework labels the respective person as abnormal. The same can happen in the case of fire or explosion, if there is a person nearby that runs away from the fire or that is thrown on the ground by the blast. A pure object-centric framework is expected to increase the number of false negatives due to detection failures, but, in the same time, it significantly reduces the number of false positives (as the framework is focused on objects). Our results show that the object-centric pipeline attains significantly better results compared to its frame-level counterpart. Thus, the benefits of the object detector outweigh its limitations. Moreover, our final framework combines both object-centric and frame-level streams, alleviating the limitations of a pure object-centric method and improving the overall performance. Indeed, the frame-level pipeline can detect all anomaly types. The frame-level framework can localize anomalies by considering the magnitude of reconstruction errors in the output of the middle frame prediction head, just as other reconstruction-based approaches.

\paragraph{Generating object-centric temporal sequences.}
We take the bounding box of an object $x$ in frame $i$ and apply the same bounding box in preceding or subsequent frames to form an object-centric temporal sequence. If the object $x$ is detected in another frame, say $i\!+\!1$, we will use the respective bounding box to generate another object-centric temporal sequence. Although we may end up with multiple slightly different sequences for the same object, this is better than applying an object tracker (which increases time and introduces errors). 

\paragraph{Notes on the chosen proxy tasks.}
We underline that anomalies can be caused by both abnormal motion and abnormal appearance. Our multi-task framework can detect both anomaly types, since the first two proxy tasks (arrow of time, motion irregularity) focus on motion anomalies, while the last two tasks (middle box prediction, knowledge distillation) focus on appearance anomalies. Although our framework is simple, it is based on careful design thinking and significant effort in formulating the proxy tasks, in a single architecture, to be beneficial for anomaly detection. We believe that its simplicity coupled with its effectiveness in anomaly detection is interesting and compelling. Nevertheless, in future work, additional or alternative proxy tasks can be considered while seeking to further improve the results.

\end{document}